\pdfoutput=1 
\documentclass[twoside,leqno,twocolumn]{article}
\usepackage[letterpaper]{geometry}
\usepackage{ltexpprt}
\usepackage[numbers, compress]{natbib}

\usepackage[utf8]{inputenc} 
\usepackage[T1]{fontenc}    
\usepackage{hyperref}       
\usepackage{url}            
\usepackage{booktabs}       
\usepackage{nicefrac}       
\usepackage{microtype}      
\usepackage{nccmath}
\usepackage{amsmath}
\usepackage{amssymb}
\usepackage{bm}
\usepackage{subcaption}
\usepackage{paralist}
\usepackage{tabularx}
\usepackage{mathtools}
\usepackage{cleveref}
\usepackage{enumitem}
\usepackage{thmtools, thm-restate}
\usepackage{multirow}
\usepackage{changepage}
\usepackage{adjustbox}
\usepackage{makecell}
\usepackage{lipsum}
\usepackage{algorithm2e}
\usepackage{algorithmic}
\usepackage{tikz}
\usepackage{dblfloatfix}
\usetikzlibrary{bayesnet}
\usepackage{pifont}
\usepackage[page]{appendix}

\DeclareMathOperator*{\argmin}{arg\,min}
\newcommand*\samethanks[1][\value{footnote}]{\footnotemark[#1]}

\newcommand{\tuple}[1]{\ensuremath{\left \langle #1 \right \rangle }}
\newcommand{\UTXTY}{\langle X,T,Y \rangle}
\newcommand{\VXTY}{X \rightarrow T \leftarrow Y}
\newcommand{\Z}{\mathbf{Z}}
\newcommand{\A}{\mathbf{A}}
\newcommand{\B}{\mathbf{B}}
\newcommand{\Zcal}{\mathcal{Z}}
\newcommand{\Scal}{\mathcal{S}}
\newcommand{\olp}{\textsc{olp}}
\newcommand{\true}{\textsc{true}}
\newcommand{\false}{\textsc{false}}
\newcommand{\bnlearn}{\href{https://www.bnlearn.com/bnrepository/}{bnlearn}}
\newcommand*\circled[1]{\tikz[baseline=(char.base)]{
            \node[shape=circle,draw,inner sep=2pt] (char) {#1};}}
\newcommand{\cmark}{\ding{51}}%
\newcommand{\xmark}{\ding{55}}%

\makeatletter
\newcommand\notsotiny{\@setfontsize\notsotiny{6.5}{8.0}}
\makeatother

\begin{document}
\title{\Large ML4C: Seeing Causality Through Latent Vicinity}
\author{%
  Haoyue Dai\thanks{Carnegie Mellon University. \href{mailto:hyda@cmu.edu}{\texttt{hyda@cmu.edu}}} \thanks{Work done during an internship at Microsoft Research Asia.}
  \and Rui Ding\thanks{Microsoft Research Asia. \href{mailto:juding@microsoft.com}{\texttt{juding@microsoft.com}}} \thanks{Corresponding author.}
  \and Yuanyuan Jiang\thanks{Renmin University of China.} \samethanks[2]
  \and Shi Han\samethanks[3]
  \and Dongmei Zhang\samethanks[3]
}
\date{}
\maketitle
\fancyfoot[R]{\scriptsize{Copyright \textcopyright\ 2023 by SIAM\\
Unauthorized reproduction of this article is prohibited}}

\begin{abstract}
  \small Supervised Causal Learning (SCL) aims to learn causal relations from observational data by accessing previously seen datasets associated with ground truth causal relations. This paper presents a first attempt at addressing a fundamental question: \textit{What are the benefits from supervision and how does it benefit?} Starting from seeing that SCL is not better than random guessing if the learning target is non-identifiable a priori, we propose a two-phase paradigm for SCL by explicitly considering structure identifiability. Following this paradigm, we tackle the problem of SCL on discrete data and propose ML4C. The core of ML4C is a binary classifier with a novel learning target: it classifies whether an Unshielded Triple (UT) is a v-structure or not. Specifically, starting from an input dataset with the corresponding skeleton provided, ML4C orients each UT once it is classified as a v-structure. These v-structures are together used to construct the final output. To address the fundamental question of SCL, we propose a principled method for ML4C featurization: we exploit the vicinity of a given UT (i.e., the neighbors of UT in the skeleton), and derive features by considering the conditional dependencies and structural entanglement within the vicinity. We further prove that ML4C is asymptotically correct. Thorough experiments conducted on benchmark datasets demonstrate that ML4C remarkably outperforms other state-of-the-art algorithms in terms of accuracy, reliability, robustness and tolerance. In summary, ML4C shows promising results on validating the effectiveness of supervision for causal learning. Our codes are publicly available at \href{https://github.com/microsoft/ML4C}{https://github.com/microsoft/ML4C}.

\end{abstract}

\section{Introduction}
\label{sec:introduction}
The problem of causal learning is to learn causal relations from observational data~\citep{glymour2019review}. The learned causal relations are typically represented in the form of a Directed Acyclic Graph (DAG), where each edge in the DAG indicates direct cause-effect relation between the parent node and child node. 

The methods of causal learning mostly fall into four categories: constraint-based, score-based, continuous optimization method and functional causal models. Each of these methods takes a given dataset as input and outputs a DAG but with different criteria. For instance, the DAG should be consistent with conditional independencies in the data (constraint-based); or it is optimal w.r.t. a pre-defined score function under either combinatorial constraint (score-based) or continuous equality constraint (continuous optimization). In a nutshell, these methods can be viewed as \textit{unsupervised} since they do not access additional datasets associated with ground truth causal relations.

A new line of research called \textit{Supervised} Causal Learning (SCL), on the other hand, aims to learn causal relations in the supervised fashion: the algorithm has access to datasets associated with ground truth causal relations, in the hope that learning causal relations on newly unseen datasets benefits from such supervision. Despite several existing works on this direction (see Related Work), a fundamental question remains unanswered: \textit{How is supervised causal learning possible}? Specifically, compared with unsupervised causal learning methods, can we gain additional benefits from supervision? If the answer is yes, then what are the benefits?

We tackle the problem by first seeing crucial connections between SCL and causal structure identifiability. Considering the problem of causal learning on discrete data, the theorem in~\citep{meek2013strong} states that, under standard assumptions (i.e., Markov assumption, faithfulness and causal sufficiency), we can only identify a graph up to its Markov equivalence class. Markov equivalence class is the set of DAGs having same skeleton and same v-structures, which can be represented by CPDAG (Completed Partially Directed Acyclic Graph). Thus, the (un)directed edges in the CPDAG indicate (non-)identifiable causal relations. Each non-identifiable edge in CPDAG can be oriented by either direction to equivalently fit the observational data. Given an SCL algorithm with the learning target as the orientation of an edge, we see that it is not better than random guessing (or could be worse due to sample bias in training data) to predict any non-identifiable edge since we can assign either $X\rightarrow Y$ or $X\leftarrow Y$ with the same input dataset.

\begin{restatable}{observation}{IDENRANDGUESS}
\label{prop:iden_rand_guess}
Considering the learning target is the orientation of an edge. If the edge is non-identifiable a priori, then supervised causal learning is no better than random guessing.
\end{restatable}

\begin{figure}[!b]
\vspace{-2em}
\includegraphics[width=1\linewidth, ]{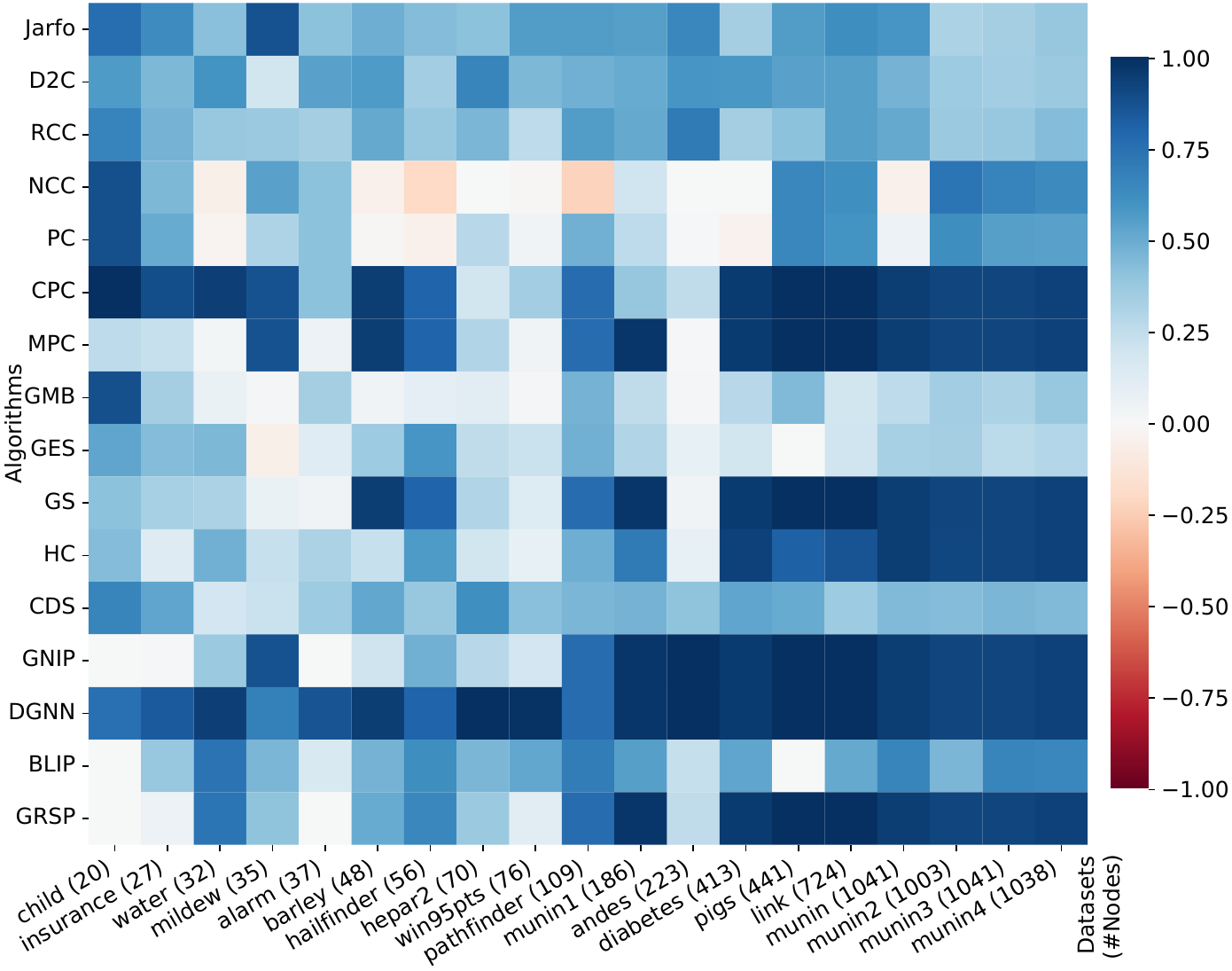}
\vspace{-1.5em}
\caption{Comparison of ML4C and other algorithms (difference between ML4C's and others' F1 scores) on benchmark datasets. Blue indicates that ML4C is superior whereas red indicates that the competitor is better.\looseness=-1}
\label{fig:evaluation_heatmap}
\end{figure}

\begin{figure*}
\begin{minipage}{0.2\textwidth}
\includegraphics[width=1\linewidth, ]{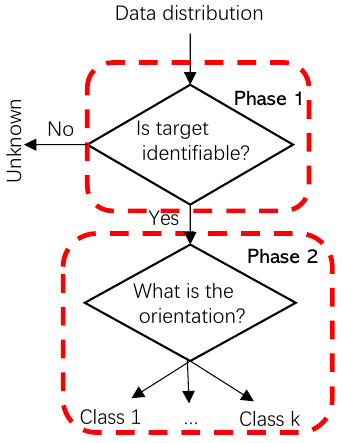}
\end{minipage}
\hfill\vline\hfill
\noindent\begin{minipage}{0.75\textwidth}
\includegraphics[width=1\linewidth, ]{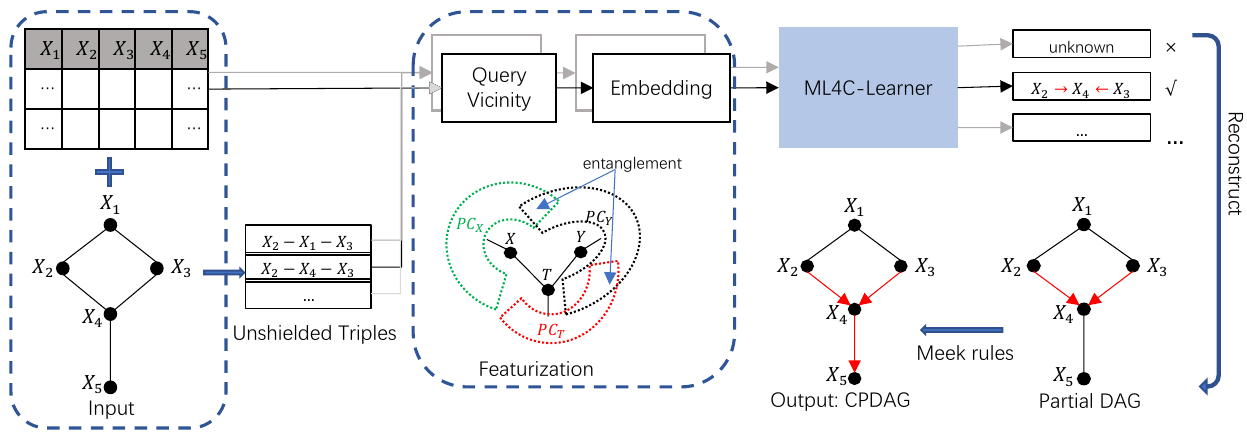}
\end{minipage}
\quad\begin{minipage}{0.2\textwidth}
\centering
\small{(a)}
\end{minipage}
\noindent\begin{minipage}{0.75\textwidth}
\centering
\small{(b)}
\end{minipage}
\centering
\caption{(a) Two-phase paradigm for supervised causal learning. (b) ML4C's workflow.}
\vspace{-1.5em}
\label{fig:two_phase_workflow}
\end{figure*}

All proofs of observations, propositions, lemmas and theorems are available in the supplementary material. Consequently, we propose and advocate a two-phase paradigm for SCL, as depicted in Figure~\ref{fig:two_phase_workflow}(a): phase one corresponds to determining the identifiability of a specific learning target; only if it is determined as identifiable, we go to phase two to classify the specific orientation of the learning target. Following this paradigm, we tackle the problem of SCL on discrete data and propose an algorithm ML4C. The core of ML4C is a binary classifier with a novel learning target: it classifies whether an Unshielded Triple (UT: a triple of variables \tuple{X,T,Y} where $X$ and $Y$ are adjacent to $T$ but are not adjacent to each other) is a v-structure or not. Starting from an input dataset with the corresponding skeleton provided, ML4C orients each UT once it is classified as a v-structure. These v-structures are further used to construct a CPDAG as output. Such a single classifier facilitates both learning tasks in the two phases, since standard theory shows that an identifiable UT implies that it is a v-structure~\citep{spirtes2000causation}, i.e., up to the partial DAG before applying Meek rules~\citep{meek2013rule}.


Operationally, we propose a principled method for ML4C featurization. We exploit the \textit{vicinity} of a given UT (i.e., the neighbors of UT in the skeleton), and derive features by considering the conditional \textit{dependencies} and structural \textit{entanglement} within the vicinity. ML4C's featurization is appealing from both theoretical and empirical perspectives. We prove that ML4C is asymptotically correct, and ML4C shows remarkable performance on benchmark datasets (with finite samples).\looseness=-1 

\vspace{0.5em}
\noindent\textbf{Evaluation Highlight}: We compare ML4C and other SOTA algorithms on the bnlearn benchmark~\citep{scutari2012bayesian} thoroughly (as shown in Figure \ref{fig:evaluation_heatmap}). Overall, ML4C significantly outperforms all competitors, with the highest average F1-score and consistent performance across all datasets (i.e., most of the blocks in Figure \ref{fig:evaluation_heatmap} are blue-colored). Also, ML4C shows high accuracy (F1-score$>0.9$) on very large-scale datasets ($>1000$ nodes) while max(others)$\sim 0.6$ (last three rows of `\texttt{munin*}' in Table \ref{table:accuracy}). Our main contributions are:\looseness=-1

\begin{enumerate}[leftmargin=0.5em,labelwidth=\itemindent,align=left]
   \item ML4C is a supervised approach for causal learning, and to the best of our knowledge, it is the first such approach to tackle discrete data systematically.\looseness=-1 
   \item ML4C is with the following novelties: \begin{inparaenum}[i)]\item\textbf{Learning Target}: The core of ML4C is a binary classifier with the orientation of a UT as its learning target to address the two-phase tasks simultaneously. \item\textbf{Featurization}: A principled method to exploit vicinity information in terms of dependencies and entanglement of a given UT. \item\textbf{Learnability}: We prove that ML4C is asymptotically correct. \item\textbf{Empirical Performance}: Experiments conducted on benchmark datasets demonstrate that ML4C remarkably outperforms other SOTAs.\end{inparaenum}\looseness=-1
\end{enumerate}


\section{Related Work}
\label{sec:related_work}

We divide literature on causal learning into supervised and unsupervised approaches, depending on whether additional datasets (associated with ground truth causal relations) are accessed (supervised) or not (unsupervised). In the literature of unsupervised causal learning, constraint-based methods aim to identify a DAG which is consistent with conditional independencies. The learning procedure of constraint-based methods first identifies the corresponding skeleton and then conducts orientation based on v-structure identification~\citep{yu2016review}. The typical algorithm is PC~\citep{PC}, and there are also PC-derived algorithms such as Conservative-PC~\citep{ramsey2012conservative} and PC-stable~\citep{colombo2014stable} which improve the robustness of v-structure identification. Score-based methods aim to find the DAG which is optimal w.r.t. a pre-defined score function under combinatorial constraints by a specific search procedure, such as forward-backward search GES~\citep{chickering2002optimal}, hill-climbing~\citep{koller2009probabilistic}, integer programming~\citep{cussens2012bayesian}, or by order search~\citep{scanagatta2015learning,lam2022greedy}. Continuous optimization methods transform the discrete search procedure into continuous equality constraint: NOTEARS~\citep{zheng2018dags} formulates the acyclic constraint as a continuous equality constraint, it is further extended by DAG-GNN~\citep{yu2019dag} and MCSL~\citep{ng2022masked} to support learning non-linear causal relations. \citep{lorch2021dibs} proposes to estimate the posterior distribution of causal DAGs.

SCL emerges from the task of orienting edge in the continuous, non-linear bivariate case under Functional Causal Model (FCM) formalism. Given a collection of cause-effect samples (dataset $\sim$ binary label indicating whether $X\rightarrow Y$ or $X\leftarrow Y$), supervised approaches such as RCC~\citep{lopez2015randomized, lopezpaz2015towards}, NCC~\citep{lopez2017discovering}, D2C~\citep{d2c2015dependency} and Jarfo~\citep{fonollosa2019conditional} achieve better performance on predicting pairwise relations (i.e., the orientation of an edge) than unsupervised approaches such as ANM~\citep{hoyer2008nonlinear} or IGCI~\citep{janzing2012information}. Differently, ~\citep{li2020supervised} sets the learning target as the whole DAG structure instead of pairwise  relation and it is applied on data which is generated by the linear Structural Equation Model (SEM). We summarize the differences in the problem space between ML4C and the other SCL approaches as follows: \begin{inparaenum}[i)]\item We advocate a two-phase learning paradigm and emphasize the relationship between identifiability and learnability. Specifically, presuming additive noise model~\citep{hoyer2008nonlinear} or linear SEM with non-Gaussian noise~\citep{shimizu2006linear} provides license to identifiability thus the aforementioned approaches can be viewed as specific tasks in phase two. \item We tackle SCL's learnability not only via empirical evaluation but also by theoretical analysis to shed light on the fundamental question of learnability. \item ML4C deals with discrete data systematically while other approaches mainly focus on continuous data.\end{inparaenum}\looseness=-1

\vspace{-0.5em}
\section{Background}
\label{sec:background}
\subsection{Basic Notations}
\label{sec:background_basic_notations}
A discrete dataset $D_i$ consists of $n_i$ records and $d_i$ categorical columns, which represents $n_i$ instances drawn i.i.d. from $d_i$ discrete variables $X_{1},X_{2}, \cdots ,X_{d_{i}}$ by a joint probability distribution $P_i$, which is entailed by an underlying data generating process, denoted as DAG $G_i$.

\noindent\textbf{Causal sufficiency}: There are no latent common causes of any of the variables in the graph.

\noindent\textbf{Markov factorization property}: Given a joint probability distribution $P$ and a DAG $G$, $P$ is said to satisfy Markov factorization property w.r.t. $G$ if $P:=P \left( X_{1},X_{2}, \cdots ,X_{d} \right) = \prod_{i=1}^{d}P \left( X_{i} \vert \text{pa}_{i}^{G} \right)$, where $ \text{pa}_{i}^{G} $  is the parent set of  $ X_{i} $  in  $ G $.

\noindent\textbf{Global Markov Property (GMP)}: $P$ is said to satisfy GMP (or Markovian) w.r.t. a DAG $G$ if $X\bot_{G}Y \vert Z \Rightarrow X\bot Y \vert Z$. Here $\bot_{G}$ denotes d-separation, and $\bot$ denotes statistical independence. GMP indicates that any d-separation in graph $G$ implies conditional independence in distribution $P$. GMP is equivalent to the Markov factorization property~\citep{lauritzen1996graphical}.

\noindent\textbf{Faithfulness}: Distribution $P$ is faithful w.r.t. a DAG $G$ if $X\bot Y \vert Z \Rightarrow X\bot_{G}Y \vert Z$.

\noindent\textbf{Canonical dataset}: We say a discrete dataset $D$ is canonical if its underlying probability distribution $P$ is Markovian and faithful w.r.t. some DAG $G$. 

\vspace{-0.8em}
\subsection{Causal Structure Identifiability}
\label{sec:background_causal_structure_identifiability}
Below are the established identifiability results on discrete data~\cite{meek2013strong}.\looseness=-1 

\vspace{-0.6em}
\begin{definition}[Markov equivalence]
\label{def:markov_equivalence}
Two graphs are Markov equivalent if and only if they have the same skeleton and same v-structures. A Markov equivalence class can be represented by a CPDAG having both directed and undirected edges. A CPDAG can be derived from a DAG $G$, denoted as CPDAG$(G)$. \normalfont{The theorem of Markov completeness in ~\citep{meek2013strong} states that, under causal sufficiency, we can only identify a causal graph up to its Markov equivalence class on canonical data. Therefore, the (non-)identifiable causal relations are the (un)directed edges in the CPDAG. Formally,}
\end{definition}

\vspace{-1.5em}
\begin{definition}[Identifiability]
\label{def:identifiability}
Assuming $P$ is Markovian and faithful w.r.t. a DAG $G$ and causal sufficiency, then each (un)directed edge in CPDAG$(G)$ indicates (non-)identifiable causal relation.
\end{definition}

\vspace{-0.8em}
\subsection{ML4C Related Notations}
\label{sec:ml4c_related_notations}
\vspace{-0.8em}
\begin{definition}[Skeleton]
\label{def:skeleton}
A skeleton $E$ defined over distribution $P \left( X_{1},X_{2}, \cdots ,X_{d} \right)$ is an undirected graph such that there is an edge between $X_i$ and $X_j$ if and only if $X_i$ and $X_j$ are always dependent, i.e., $\nexists Z \subseteq  \left\{ X_{1},X_{2}, \cdots ,X_{d} \right\} ~s.t.~X_{i}\bot X_{j} \vert Z $. \normalfont{Skeleton is a statistical concept, which can be obtained prior to facilitating various downstream tasks. Recently, there have been some novel skeleton learning algorithms such as ~\citep{ding2020reliable}. In particular, skeleton can be used for causal learning: the theorem in ~\citep{spirtes2000causation} states that if distribution $P$ is Markovian and faithful w.r.t. a DAG $G$, then skeleton $E$ is the same as the undirected graph of $G$.} 
\end{definition}
\vspace{-1.0em}
\begin{definition}[UT]
\label{def:ut}
A triple of variables $\langle X,T,Y \rangle$ in a skeleton is an unshielded triple, or short for UT, if $X$ and $Y$ are adjacent to $T$ but are not adjacent to each other. $\UTXTY$ can be further oriented to become a v-structure $\VXTY$, in which $T$ is called the collider.\looseness=-1
\end{definition}
\vspace{-1.2em}
\begin{definition}[PC]
\label{def:pc}
Denote the set of parents and children of $X$ in a skeleton as $PC_X$, in other words, $PC_X$ are the neighbors of $X$ in the skeleton. For convenience, if we discuss $PC_X$ in the context of a UT $\UTXTY$, we intentionally mean the set of parents and children of $X$ but exclude $T$. Similarly, $PC_T$ excludes $X,Y$.
\end{definition}
\vspace{-1.2em}
\begin{definition}[Vicinity]
\label{def:vicinity}
We define the vicinity of a UT $\UTXTY$ as $ V_{ \langle X,T,Y \rangle }:= \{ X,T,Y \}  \cup PC_{X} \cup PC_{Y} \cup PC_{T} $. Vicinity is a generalized version of PC, i.e., the neighbors of $\{ X,T,Y \}$ in the skeleton.
\end{definition}
\vspace{-1.2em}
\begin{definition}[Sepsets]
\label{def:sepsets}
Denoted as $\Scal:=\left\{ S:X\bot Y \vert S, S \subset PC_{X} \cup T, \text{or} S \subset PC_{Y} \cup T \right\}$. \normalfont{Under faithfulness assumption, sepsets $\Scal$ is an ensemble where each item is a subset of variables within the vicinity that d-separates $X$ and $Y$.}
\end{definition}


\section{Approach}
\label{sec:approach}
\subsection{Viewing PC and MPC as Special Classifiers}
\label{sec:viewing_pc_mpc}
The motivation of ML4C is from the  observation, that the orientation logics of both PC~\citep{spirtes2000causation} and MPC~\citep{colombo2014stable} can be interpreted as explicit featurization and static classification mechanisms. Starting from PC, it first learns the skeleton from data, and then conducts orientation by taking skeleton as input. Specifically, for each UT $\UTXTY$ (queried from skeleton), it finds a sepset $S$ such that $X\bot Y | S$, and then simply checks if $T\in S$ or $T\not\in S$: if $T\not\in S$, orient $\UTXTY$ as a v-structure, and otherwise undetermined. From ML perspective, we re-formulate PC's logic as follows:

\noindent\textbf{Task}: To classify if a UT $\UTXTY$ is a v-structure.

\noindent\textbf{Featurization}: Finds a sepset $S$ s.t. $X\bot Y | S$, and defines a Boolean feature $x_{\texttt{PC}}(\UTXTY)\coloneqq T\in S$.

\noindent\textbf{Classifier}: $C_{\texttt{PC}}(x_{\texttt{PC}})\coloneqq \left\{ \begin{aligned}\text{non-v} \ &x_{\texttt{PC}}=\true \\ \text{v-struc} \ &x_{\texttt{PC}}=\false \end{aligned}\right.$.\looseness=-1

\noindent Majority-PC (MPC)~\citep{colombo2014stable} is a sample version enhancement of PC's orientation, which achieves better performance on finite samples. Instead of finding only one sepset $S$, MPC finds all the sepsets $\Scal$ and counts the number of sepsets $S_i\in\Scal$ that contains $T$. We summarize its logic in the ML formulae:

\noindent\textbf{Featurization}: Finds all sepsets $\Scal$ of $X,Y$, and defines a real-value feature $x_{\texttt{MPC}}(\UTXTY)\coloneqq \frac{|\{S_i | T\in S_i \in \Scal\}|}{|\Scal|}$.

\noindent\textbf{Classifier}: $C_{\texttt{MPC}}(x_{\texttt{MPC}})\coloneqq \left\{ \begin{aligned}\text{non-v} \ &x_{\texttt{MPC}}>0.5 \\ \text{v-struc} \ &x_{\texttt{MPC}}\leq 0.5 \end{aligned}\right.$.\looseness=-1

\begin{observation}
Both the ``hand-crafted'' classifiers of PC and MPC are asymptotically correct. Furthermore, $C_{\texttt{MPC}}$ is more ``sophisticated'' than $C_{\texttt{PC}}$ due to its more complex classification mechanism, and thus $C_{\texttt{MPC}}$ achieves better performance empirically. However, from the ML perspective, either $C_{\texttt{PC}}$ or $C_{\texttt{MPC}}$ is simple, so we are encouraged to provide more systematic featurization, and to learn a better classification mechanism.
\end{observation}

\vspace{-1.2em}
\subsection{ML4C's Overview}
\label{sec:overview}
Figure \ref{fig:two_phase_workflow}(b) depicts the overall workflow of ML4C, which is composed of ML4C-Learner with other standard logics. Similar to~\citep{spirtes2000causation}, ML4C-Learner classifies whether a UT is a v-structure or not. Specifically, featurization is conducted to represent each UT as an embedded vector, which is fed into ML4C-Learner, a binary classifier. In the inference stage, we obtain all the v-structures which are classified by ML4C-Learner and reconstruct a partial DAG and then, a CPDAG is output by applying Meek rules on the partial DAG. In the training stage, the label of each UT is obtained by querying from ground truth DAG $G_{i}$. We collect labeled training data from synthesis.

By Markov completeness, the set of v-structures is invariant across all Markov equivalent DAGs for a canonical dataset, and it can fully recover the CPDAG, provided that the skeleton is given. Thus, besides its dedicated role in phase 2, ML4C-Learner also facilitates learning task in phase 1 since an identifiable UT implies that it is a v-structure (up to the partial DAG before applying Meek rules).
\vspace{-0.7em}
\subsection{Featurization}
\label{sec:featurization}
We propose a principled method for ML4C-Learner's featurization. We further prove that ML4C-Learner is asymptotically correct.
\vspace{0.5em}

\noindent\textbf{Design Principles}: Our key aspect of featurization is to broaden the focus from a specific UT to its \textit{vicinity} and seeking conditional \textit{dependencies} and structural \textit{entanglement} within the vicinity, to reveal reliable asymmetry that distinguishes v-strucs and non-v-strucs.\looseness=-1 
\vspace{0.5em}
\noindent$\bullet$ \textbf{\textit{Dependencies within Vicinity}}

\noindent\textbf{Conditional dependency}: Denoted as $ X \sim Y \vert \Z $, which is a non-negative scalar that measures the dependence between two random variables $X$ and $Y$ given variable set $\Z$. Operationally, $ X \sim Y \vert \Z $ is composed of two parts, bivariable $ X \sim Y $, and conditional $\Z$. We further extend the definition to allow a set of variables in bivariable, and an ensemble (i.e., a set of set) as conditional:\looseness=-1

\noindent\textbf{Extended conditional dependency}: Denoted as $\A \sim \B \vert \Zcal$$:= \{ X \sim Y \vert \Z:$$X \in \A,Y \in \B,\Z \in \Zcal \} $, where $\A$ and $\B$ are set of variables, and $\Zcal$ is an ensemble. Thus, extended conditional dependency is a set of scalars.

Within the vicinity of $\UTXTY$, we start from measuring dependencies between $\left\{ X, PC_{X} \right\}$ and $\left\{ Y, PC_{Y} \right\}$ by conditioning on $\left\{ T, PC_{T} \right\}$. Intuitively, if $\UTXTY$ is a v-structure, conditioning on $T$ or $T$'s descendants tends to strengthen the dependency between $PC_X$ and $PC_Y$ since the paths passing $X-T-Y$ are unblocked; otherwise, conditioning on $T$ tends to weaken the dependency between $PC_X$ and $PC_Y$ because $T$ blocks the paths passing $X-T-Y$. Therefore, such conditional dependencies reflect potential asymmetry to distinguish v-structure and non-v-structure. Formally, 

\begin{definition}[Domain of bivariable]
\label{def:domain_of_bivariable}
Denoted as $ \mathbb{B}:= \left\{ X, PC_{X} \right\}  \times  \left\{ Y, PC_{Y} \right\}  \equiv  \left\{ X \sim Y,X \sim PC_{Y},PC_{X} \sim Y,PC_{X} \sim PC_{Y} \right\}$, here symbol $\times$ is Cartesian product.
\end{definition}

\begin{definition}[Domain of conditional]
\label{def:domain_of_conditional}
Denoted as $\mathbb{C}:= \left\{  \varnothing ,T,\mathcal{PC}_{T} \right\}  \vee  \left\{  \varnothing ,S \right\}  \equiv  \left\{  \varnothing ,T,\mathcal{PC}_{T},\Scal,\Scal \vee T,\Scal \vee \mathcal{PC}_{T} \right\}$, where $\mathcal{PC}_{T}:= \left\{  \left\{ I \right\} : I \in PC_{T} \right\}$ which is an ensemble version of $PC_{T}$, and $\Scal \vee \mathcal{PC}_{T}:= \left\{ S \cup I:S \in \Scal,I \in \mathcal{PC}_{T} \right\}$. Here symbol $\vee$ is element-wise union.
\end{definition}
We exploit the extended conditional dependencies from $\mathbb{B} \times \mathbb{C}$, i.e., we pick a bivariable from $\mathbb{B}$ and a conditional from $\mathbb{C}$, and calculate the extended conditional dependency. There are in total $\vert \mathbb{B} \vert  \times  \vert \mathbb{C} \vert =24$ extended conditional dependencies.

\begin{restatable}{lemma}{SEPSETSNONEMPTY}
\label{lem:sepsets_nonempty}
Sepsets $\Scal$ of a UT is non-empty. 
\end{restatable}
\begin{restatable}{remark}{SEPSETSSCOPE}
\label{rem:sepsets_scope}
We intend to restrict the sepsets within the vicinity of $\UTXTY$. Lemma \ref{lem:sepsets_nonempty} shows the existence of such d-separation sets within vicinity. Furthermore, searching for all d-separation sets is highly time-consuming, thus significant computational cost is also saved.
\end{restatable}
\vspace{0.5em}

\noindent$\bullet$ \textbf{\textit{Entanglement within Vicinity}}

Structural entanglement reflects complex structure within the vicinity of $\UTXTY$. Variables $X$, $Y$ and $T$ can mutually share common neighbors, and their neighbors may also overlap with sepsets $\Scal$. We call such overlaps structural entanglement. Specifically, we exploit the overlap coefficient ~\citep{vijaymeena2016similarity} to measure the entanglement: 
\begin{definition}[Overlap coefficient]
\label{def:overlap_coefficient}
$\olp \left( \A,\B \right):=\vert \A \cap \B \vert / \min  \left(  \vert \A \vert , \vert \B \vert  \right)$, where $\A$ and $\B$ are two sets of variables. \normalfont{We extend this formula to support the ensemble as input:}
\end{definition}
\noindent\textbf{(Extended) Overlap coefficient}: $\olp \left( \A,\Scal \right) :=  \sum _{i=1}^{ \vert \Scal \vert }\olp \left( \A, S_{i} \right) / \vert \Scal \vert $. Naturally, we consider the entanglement in terms of the overlap coefficient on each pair of items in domain $\left\{ PC_{X},PC_{Y},PC_{T},\Scal \right\}$. Thus, use 7 scalars (including $\olp \left( \{T\},\Scal \right)$) to represent the entanglement within the vicinity of a UT.


\noindent\textbf{Embedding}: We aim to represent the dependencies and entanglement by a feature vector with fixed dimensionality, which can be used to train ML4C-Learner. We adopt the standard kernel mean embedding technique in~\citep{smola2007hilbert}. Details are available in Appendix (\cref{app:implementation}). 

\subsection{Learnability}
\label{sec:learnability}
We have presented ML4C's featurization and see that conditional dependencies and structural entanglement have the potential to reveal asymmetry to distinguish v-structure and non-v-structure UTs. Now we provide rigorous analysis to show that, for a canonical dataset with sufficient samples, ML4C-Learner is asymptotically correct. We first propose a surrogate object called discriminative predicate:\looseness=-1

\begin{definition}[Discriminative predicate]
\label{def:discriminative_predicate}
A discriminative predicate is a binary predicate function with domain as ML4C's feature vector. \normalfont{A discriminative predicate can be viewed as a special classifier with pre-specified form of mechanism (e.g., $C_\texttt{PC}$ or $C_\texttt{MPC}$ is a discriminative predicate).}
\end{definition}
\begin{definition}[Weak/Strong predicate]
\label{def:weak_strong_discriminative_predicate}
Whenever a discriminative predicate takes the feature vector of a UT as input, a weak (discriminative) predicate satisfies one of the following two criteria; a strong (discriminative) predicate satisfies both: \begin{inparaenum}[i)]\item it is evaluated to $\true$ if the UT is a v-structure; \item it is evaluated to $\false$ if the UT is not a v-structure.\end{inparaenum}
\end{definition}
By definition, a weak  predicate exhibits discriminative power since its evaluation as false implies the UT is a non-v-structure (or true implies v-structure). A strong predicate is sound and complete (e.g., $C_\texttt{MPC}$). Denote $\left\{ \A \sim \B \vert \Zcal \right\} > \delta :=X \sim Y \vert \Z> \delta : \forall X \in \A,Y \in \B,\Z \in \Zcal$, then we have:

\begin{restatable}[Existence of strong predicate]{lemma}{STRONGEXISTS}
\label{lem:existence_of_strong_discriminative_predicate}
For a canonical dataset with infinite samples, the following are three strong discriminative predicates: \begin{inparaenum}[i)]\item $\olp(T,\Scal)=0$, \item $\olp(T,\Scal)<0.5$, \item $\olp(T,\Scal)<1\wedge\min  \left\{ X \sim Y \vert T \cup \Scal \right\} >0$.\end{inparaenum} \looseness=-1
\end{restatable}

For example, $\textit{ii)}$ is the orientation logic of $C_{\texttt{MPC}}$.

\begin{restatable}[Existence of weak predicate]{lemma}{WEAKEXISTS}
\label{lem:existence_of_weak_discriminative_predicate}
For a canonical dataset with infinite samples, the following are three weak discriminative predicates: \begin{inparaenum}[i)]\item $\left\{ X \sim Y \vert T \right\} >0$, \item $\left\{ X \sim Y \vert \mathcal{PC}_{T} \right\} =0$ , \item $\left\{ PC_{X} \sim PC_{Y} \vert \Scal \cup T \right\} >0$.\end{inparaenum}
\end{restatable}
Take $\textit{i)}$ $\left\{ X \sim Y \vert T \right\} >0$ for example, if $\UTXTY$ is a v-structure, then $T$ is a collider that unblocks $X$ and $Y$ through path $X-T-Y$, so $\left\{ X \sim Y \vert T \right\} >0$. Similarly, $\textit{ii)}$ if a UT is not a v-structure, then $\forall pc_T \in\mathcal{PC}_T$, $X$ and $Y$ must be d-connected by $pc_T$. Weak predicates exploit graphical implications within the vicinity. They are valid in one direction, i.e., either sufficient or necessary but not both. In addition to strong predicates, such one-direction-asymmetry exhibits additional discriminative power from the ML perspective.

\begin{restatable}{theorem}{TENDPERFECT}
\label{thrm:tend_perfect}
ML4C-Learner is asymptotically correct on classifying a canonical dataset with sufficient samples.
\end{restatable}

\section{Evaluation}
\label{sec:evaluation}
\noindent\textbf{Benchmark Datasets}:
\label{sec:benchmark_datasets}
We use discrete datasets of all 24 networks from \bnlearn~repository~\citep{scutari2012bayesian} for evaluation. For each dataset we sample 1k, 5k, 10k, 15k, 20k records for use.\looseness=-1

\noindent\textbf{ML4C's Training and Inference}:
\label{sec:ml4c_training_and_inference}
We generate ML4C's training data synthetically (which is also used for other SCL competitors). Specifically, 400 unique DAGs are randomly generated by two models: Erd\H{o}s-R\'{e}nyi (ER) model~\citep{erdos1959on} and Scale-Free (SF) model~\citep{albert2002statistical}, with the number of nodes ranging from 10 to 1,000. A standard random forward data generation process is applied to obtain 10k observational samples for each graph. We further extract UTs from the 400 DAGs, consisting of 97,010 v-structures (label = 1) and 195,691 non-v-structures (label = 0). We use these instances to train ML4C-Learner, which is implemented by a XGBoost~\citep{chen2016xgboost} binary classifier with default hyper-parameters and we use binary cross-entropy as the loss function. Details on our synthesis procedure, configurations and implementation of ML4C-Learner are available in the supplementary material.

\noindent\textbf{Competitors}:
\label{sec:competitors}
We categorize state-of-the-art causal learning algorithms from two aspects, supervised vs. unsupervised, and can or cannot take a skeleton as input. We choose Jarfo~\citep{fonollosa2019conditional}, D2C~\citep{d2c2015dependency}, RCC~\citep{lopez2015randomized}, and NCC~\citep{lopez2017discovering} as SCL competitors. Same as ML4C, all these algorithms require a skeleton as input. All these algorithms use ML4C's training set for training and with appropriate learning target extracted. Regarding unsupervised algorithms, we choose PC~\citep{PC}, Conservative-PC (CPC)~\citep{ramsey2012conservative}, Majority-rule PC (MPC)~\citep{colombo2014stable}, GLL-MB (GMB)~\citep{aliferis2010local}, GES~\citep{chickering2002optimal}, Grow-Shrink (GS)~\citep{margaritis2003learning}, Hill-Climbing (HC)~\citep{koller2009probabilistic}, Conditional Distribution Similarity (CDS)~\citep{fonollosa2019conditional} and GOBNILP (GNIP)~\citep{cussens2012bayesian}. which can also take skeleton as input. Lastly, we also compare with DAG-GNN (DGNN)~\citep{yu2019dag}, BLIP~\citep{scanagatta2015learning}, and GRaSP (GRSP)~\citep{lam2022greedy}, which are unsupervised algorithms but cannot take skeletons as input. All these competitors are capable of dealing with discrete data. All experiments are done in a Windows Server with 2.8GHz Intel E5-2680 CPU and 256G RAM. Details are in the appendix (\cref{app:implementation}).\looseness=-1

\noindent\textbf{Design}:
\label{sec:design}
Our evaluation mainly consists of two parts: end-to-end comparison with competitors on benchmark datasets, and in-depth experiments on ML4C's learnability. The latter is further divided into four aspects: \begin{inparaenum}[i)]\item \textbf{Towards a perfect classifier}. ML4C-Learner is asymptotically correct, and we would like to know far it is from such a perfect classifier in real-world settings. \item \textbf{Reliability} (against weak/strong predicates). As stated in lemma~\ref{lem:existence_of_weak_discriminative_predicate} and \ref{lem:existence_of_strong_discriminative_predicate}, there exist weak and strong predicates, which provide prediction power of ML4C's feature vector. Thus, we would like to see how ML4C-Learner takes advantage of machine learning, to learn a more reliable classification mechanism than individual weak/strong predicates. \item \textbf{Robustness} (against varied sample size). It is known that many causal learning algorithms lack robustness w.r.t sample noise for finite datasets~\citep{lopez2015randomized}, especially CI tests are error-prone on small samples for constraint-based algorithms. We want to evaluate the robustness of ML4C (i.e., the classification mechanism) against varied sample sizes. \item \textbf{Tolerance} (to imperfect skeleton input). Besides using ground-truth skeleton as input, it is also interesting to see how ML4C's performance changes when the input skeleton is misspecified. This also provides a fairer comparison with the baseline approaches which do not take skeleton as input.\end{inparaenum}\looseness=-1

\begin{table*}[t]
  \scriptsize
   \setlength{\tabcolsep}{3.0pt}
   \renewcommand{\arraystretch}{0.4}
  \caption{Experiment results for end-to-end comparison with SOTA causal learning algorithms on benchmark datasets. Algorithm names are abbreviated. SHD and F1-score are reported. The last two rows show statistics of rank by SHD and F1-score for all competitors (Note: F1-score is at UT level).}
  \label{table:accuracy}
\begin{center}
\vspace{-1em}
  \begin{tabular}{c c c c c c c c c c c c c c c c c c c}
    \toprule
    \multicolumn{2}{c}{\multirow{2}{*}{\makecell{Datasets\\ \#nodes/\#edges}}} &
    \multicolumn{5}{c}{supervised} &
    \multicolumn{9}{c}{unsupervised} &
    \multicolumn{3}{c}{no skeleton input} \\
    \cmidrule(r){3-7} \cmidrule(r){8-16} \cmidrule(r){17-19} & 
      & ML4C  &  Jarfo & D2C & RCC  & NCC  &  PC  & CPC  & MPC  &  GMB  & GES & GS & HC & CDS & GNIP & DGNN & BLIP & GRSP\\
    \midrule

child & SHD & \textbf{0} & 18 & 16 & 18 & 20 & 22 & 13 & 9 & 20 & 15 & 13 & 13 & 18 & \textbf{0} & 23 & \textbf{0} & \textbf{0} \\
20/25 & F1 & \textbf{ 1.0} &  .24 &  .43 &  .33 &  .12 &  .12 &  .00 &  .74 &  .12 &  .47 &  .59 &  .57 &  .34 & \textbf{ 1.0} &  .25 & \textbf{ 1.0} & \textbf{ 1.0} \\
\cmidrule(r){1-19}
insurance & SHD & \textbf{5} & 41 & 30 & 34 & 28 & 36 & 34 & 21 & 29 & 34 & 28 & 19 & 36 & 11 & 53 & 35 & 14 \\
27/52 & F1 & \textbf{ .89} &  .26 &  .44 &  .42 &  .44 &  .39 &  .00 &  .66 &  .55 &  .46 &  .56 &  .76 &  .36 &  .88 &  .05 &  .51 &  .83 \\
\cmidrule(r){1-19}
water & SHD & 5 & 33 & 43 & 31 & \textbf{0} & 4 & 60 & 7 & 8 & 38 & 27 & 38 & 18 & 34 & 61 & 65 & 58 \\
32/66 & F1 &  .94 &  .52 &  .34 &  .56 & \textbf{ 1.0} &  .97 &  .00 &  .91 &  .87 &  .49 &  .62 &  .46 &  .76 &  .57 &  .00 &  .20 &  .21 \\
\cmidrule(r){1-19}
mildew & SHD & 6 & - & 17 & 25 & 34 & 21 & - & - & 7 & \textbf{3} & 9 & 23 & 18 & - & 52 & 36 & 37 \\
35/46 & F1 &  .87 & - &  .68 &  .50 &  .33 &  .56 & - & - &  .85 & \textbf{ .93} &  .80 &  .64 &  .65 & - &  .19 &  .41 &  .47 \\
\cmidrule(r){1-19}
alarm & SHD & \textbf{1} & 21 & 26 & 18 & 20 & 20 & 20 & 6 & 17 & 8 & 3 & 21 & 18 & 2 & 46 & 17 & 2 \\
37/46 & F1 & \textbf{ .98} &  .57 &  .44 &  .64 &  .57 &  .57 &  .57 &  .92 &  .64 &  .86 &  .94 &  .66 &  .62 & \textbf{ .98} &  .12 &  .82 & \textbf{ .98} \\
\cmidrule(r){1-19}
barley & SHD & 5 & 48 & 55 & 50 & \textbf{0} & 3 & - & - & 8 & 42 & - & 34 & 50 & 34 & 87 & 60 & 70 \\
48/84 & F1 &  .95 &  .46 &  .38 &  .44 & \textbf{ 1.0} &  .96 & - & - &  .91 &  .59 & - &  .72 &  .43 &  .74 &  .00 &  .48 &  .45 \\
\cmidrule(r){1-19}
hailfinder & SHD & 11 & 47 & 41 & 43 & \textbf{0} & 17 & - & - & 26 & 60 & - & 59 & 44 & 56 & 76 & 111 & 114 \\
56/66 & F1 &  .80 &  .37 &  .45 &  .42 & \textbf{ 1.0} &  .85 & - & - &  .70 &  .21 & - &  .23 &  .42 &  .32 &  .00 &  .18 &  .15 \\
\cmidrule(r){1-19}
hepar2 & SHD & \textbf{0} & 54 & 81 & 59 & \textbf{0} & 35 & 27 & 37 & 14 & 46 & 40 & 35 & 75 & 48 & 123 & 79 & 64 \\
70/123 & F1 & \textbf{ 1.0} &  .59 &  .34 &  .54 & \textbf{ 1.0} &  .72 &  .81 &  .70 &  .89 &  .75 &  .70 &  .81 &  .39 &  .72 &  .00 &  .54 &  .63 \\
\cmidrule(r){1-19}
win95pts & SHD & 1 & 65 & 51 & 33 & \textbf{0} & 8 & 42 & 7 & 5 & 32 & 21 & 16 & 50 & 27 & 112 & 103 & 23 \\
76/112 & F1 &  .99 &  .43 &  .54 &  .73 & \textbf{ 1.0} &  .95 &  .64 &  .95 &  .97 &  .77 &  .85 &  .91 &  .57 &  .81 &  .00 &  .47 &  .88 \\
\cmidrule(r){1-19}
pathfinder & SHD & 25 & 157 & 145 & 151 & \textbf{0} & 150 & - & - & 147 & 158 & - & 168 & 148 & 119 & 196 & 241 & - \\
109/195 & F1 &  .77 &  .21 &  .29 &  .21 & \textbf{ 1.0} &  .29 & - & - &  .30 &  .29 & - &  .28 &  .31 &  .00 &  .00 &  .07 & - \\
\cmidrule(r){1-19}
munin1 & SHD & \textbf{10} & 169 & 154 & 153 & 72 & 86 & 117 & - & 84 & 109 & - & 233 & 151 & 265 & - & 257 & - \\
186/273 & F1 & \textbf{ .97} &  .42 &  .47 &  .46 &  .77 &  .71 &  .58 & - &  .72 &  .67 & - &  .26 &  .50 &  .00 & - &  .42 & - \\
\cmidrule(r){1-19}
andes & SHD & \textbf{0} & 226 & 209 & 246 & \textbf{0} & 4 & 83 & 4 & 5 & 47 & 15 & 38 & 149 & 328 & - & 175 & 146 \\
223/338 & F1 & \textbf{ 1.0} &  .35 &  .41 &  .29 & \textbf{ 1.0} &  .99 &  .75 &  .99 &  .98 &  .92 &  .96 &  .92 &  .60 &  .00 & - &  .76 &  .75 \\
\cmidrule(r){1-19}
diabetes & SHD & 25 & 220 & 395 & 237 & 48 & \textbf{0} & - & - & 204 & 146 & - & 592 & 368 & - & - & 534 & - \\
413/602 & F1 &  .96 &  .62 &  .38 &  .62 &  .96 & \textbf{ 1.0} & - & - &  .68 &  .77 & - &  .03 &  .43 & - & - &  .43 & - \\
\cmidrule(r){1-19}
pigs & SHD & \textbf{0} & 350 & 332 & 263 & 400 & 400 & - & - & 268 & \textbf{0} & - & 532 & 316 & - & - & 6 & - \\
441/592 & F1 & \textbf{ 1.0} &  .44 &  .46 &  .59 &  .35 &  .35 & - & - &  .56 & \textbf{ 1.0} & - &  .18 &  .50 & - & - & \textbf{ 1.0} & - \\
\cmidrule(r){1-19}
link & SHD & \textbf{0} & 731 & 630 & 638 & 749 & 737 & - & - & 204 & 324 & - & 1047 & 400 & - & - & 947 & - \\
724/1125 & F1 & \textbf{ 1.0} &  .38 &  .45 &  .45 &  .39 &  .40 & - & - &  .81 &  .80 & - &  .14 &  .64 & - & - &  .49 & - \\
\cmidrule(r){1-19}
munin & SHD & 72 & 967 & 790 & 816 & \textbf{0} & 156 & - & - & 458 & 661 & - & 1397 & 795 & - & - & 1599 & - \\
1041/1397 & F1 &  .95 &  .36 &  .48 &  .44 & \textbf{ 1.0} &  .89 & - & - &  .69 &  .62 & - &  .00 &  .51 & - & - &  .29 & - \\
\cmidrule(r){1-19}
munin2 & SHD & \textbf{118} & 554 & 611 & 646 & 1052 & 898 & - & - & 536 & 632 & - & 1240 & 753 & - & - & 1321 & - \\
1003/1244 & F1 & \textbf{ .92} &  .60 &  .56 &  .55 &  .19 &  .30 & - & - &  .57 &  .58 & - &  .01 &  .49 & - & - &  .46 & - \\
\cmidrule(r){1-19}
munin3 & SHD & \textbf{113} & 616 & 629 & 688 & 1048 & 860 & - & - & 544 & 566 & - & 1306 & 819 & - & - & 1539 & - \\
1041/1306 & F1 & \textbf{ .92} &  .58 &  .57 &  .54 &  .25 &  .37 & - & - &  .60 &  .65 & - &  .00 &  .46 & - & - &  .26 & - \\
\cmidrule(r){1-19}
munin4 & SHD & \textbf{126} & 696 & 658 & 776 & 1058 & 876 & - & - & 649 & 618 & - & 1388 & 812 & - & - & 1627 & - \\
1038/1388 & F1 & \textbf{ .93} &  .54 &  .56 &  .50 &  .29 &  .39 & - & - &  .55 &  .64 & - &  .00 &  .49 & - & - &  .28 & - \\

\midrule
\multirow{2}{*}{rank(SHD)} & mean & \textbf{1.5} & 9.5 & 8.7 & 8.4 & 5.2 & 6.5 & 11.3 & 9.9 & 4.6 & 6.3 & 10.0 & 9.0 & 8.4 & 9.4 & 14.1 & 11.0 & 10.6 \\
 & $\pm$stdd & \textbf{0.7} & 3.7 & 4.0 & 2.6 & 4.4 & 4.0 & 3.0 & 4.2 & 2.7 & 3.3 & 3.6 & 2.7 & 2.8 & 4.4 & 2.2 & 3.8 & 4.0 \\
\midrule
\multirow{2}{*}{UT-F1} & mean & \textbf{.90} & .22 & .19 & .27 & .66 & .50 & .53 & .87 & .59 & .54 & .77 & .47 & .30 & .55 & .09 & .36 & .55 \\
 & $\pm$stdd & \textbf{.13} & .17 & .13 & .18 & .40 & .34 & .33 & .16 & .32 & .28 & .24 & .35 & .22 & .42 & .07 & .29 & .38 \\
    \bottomrule
  \end{tabular} 
  \end{center}
  \vspace{-2em}
\end{table*}
\begin{table}[t]
  \tiny
   \setlength{\tabcolsep}{1.40pt}
   \caption{Reliability: average F1-score of ML4C vs. 8 discriminative predicates derived from ML4C features on benchmarks.}
  \label{table:reliability}
\begin{center}
\vspace{-1em}
  \begin{tabular}{c c c c c c c c c c}
    \toprule&&
    \multicolumn{4}{c}{strong predicates} &
    \multicolumn{4}{c}{weak predicates} \\
     \cmidrule(r){3-6} \cmidrule(r){7-10} 
    id & ML4C &  1 & 2 & 3  & 4  &  1  & 2  & 3  & 4\\
    \midrule
 F1& \textbf{.92$\pm$.20} & .77$\pm$.31 & .52$\pm$.27 & .38$\pm$.25 & .66$\pm$.27 & .72$\pm$.25 & .61$\pm$.29 & .73$\pm$.30 & .55$\pm$.27 \\
    \bottomrule
  \end{tabular}
  \end{center}
  \vspace{-2em}
\end{table}

\vspace{0.5em}
\noindent\textbf{Metrics}:
\label{sec:metrics}
We use two standard metrics for performance evaluation: Structural Hamming Distance (SHD) and F1-score. For each dataset, we measure the SHD / F1-score of the output CPDAG (learned by a specific algorithm) against the ground truth CPDAG. Specifically, SHD is calculated at CPDAG level, which is the smallest number of edge additions, deletions, direction reversals and type changes (directed vs. undirected) to convert the output CPDAG to ground truth CPDAG. F1-score is calculated over identifiable edges. Roughly, F1-score can be viewed as a normalized version of SHD. Now we present the experiment results:\looseness=-1

\vspace{0.7em}

\noindent\textbf{End-to-End Comparison}:
\label{sec:end_to_end_comparison}Due to page limits, we report SHD and F1-score of all algorithms on top-19 large-scale datasets (results of the other 5 smallest datasets are omitted due to triviality), as depicted in Table~\ref{table:accuracy}. `-' means the algorithm fails on the dataset (either out-of-memory/exceeds 24 hours execution time/break caused by unknown errors). ML4C significantly outperforms all other competitors. The average F1-score of ML4C is the highest (0.92, first column in Table~\ref{table:reliability}). Moreover, ML4C exhibits the most stable performance across all datasets, its average ranking is $1.5\pm0.7$, while the second best is GLL-MB (GMB), with average ranking $4.4\pm2.4$. NCC is the strongest SCL competitor. It performs well on some datasets but overall its performance fluctuates significantly. Overall it only ranks $5.1\pm4.2$. Last but not least, ML4C shows high accuracy (F1$>0.9$) on very large-scale datasets (e.g., medicine datasets `munin2/3/4'~\citep{andreassen1987munin}) while $\max(\text{others})\sim 0.6$.\looseness=-1

\vspace{0.3em}

\noindent\textbf{Towards a Perfect Classifier}:
\label{sec:towards_a_perfect_classifier}
The last row of Table~\ref{table:accuracy} shows the performance of ML4C-Learner component at UT level by UT-F1 (i.e., F1-score of classifying UTs): such UT level accuracy is crucial for causal learning on discrete data, since the set of v-structures is invariant across all Markov equivalent DAGs and it can fully recover the CPDAG. The average F1-score of ML4C-Learner is $0.90\pm0.13$, which shows promising results towards a perfect classifier.

\vspace{0.5em}
\noindent\textbf{Reliability}:
\label{sec:reliability}
We manually identify 4 strong predicates and 4 weak predicates and treat each one as a replacement of ML4C-Learner. Table~\ref{table:reliability} shows the performance of these predicates. Although most predicates show value on discriminating UTs (e.g., 5/8 predicates are with $>0.6$ F1-score), ML4C-Learner has higher performance (average F1-score = 0.92) than each individual predicate (best average F1-score = 0.77). Thus, it is evident that ML4C-Learner learns a more reliable classification mechanism, by taking advantage of machine learning techniques.

\vspace{0.5em}

\noindent\textbf{Robustness}:
\label{sec:robustness}
To evaluate robustness, ML4C is trained on synthetic datasets with sample size = 10k, but it is tested on benchmark datasets with different sample sizes: 1k, 5k, 10k, 15k and 20k respectively. Table~\ref{table:robustness} shows that ML4C exhibits satisfactory robustness (decrease of F1-score is less than 0.1) against sample size on 17 out of 18 datasets.
\begin{table}[t]
  \tiny
   \setlength{\tabcolsep}{0.5pt}
   \renewcommand{\arraystretch}{0.6}
  \caption{Robustness: ML4C is trained on synthetic datasets with sample size = 10k, but tested on benchmark datasets with different sample sizes $\in \{\text{1k, 5k, 10k, 15k, 20k}\}$.}
  \label{table:robustness}
\begin{center}
\vspace{-2em}
  \begin{tabular}{c || c c c c c c | c c c c c c | c c c c c c }
    \toprule
     & size & 1k & 5k & 10k & 15k & 20k & size & 1k & 5k & 10k & 15k & 20k & size & 1k & 5k & 10k & 15k & 20k \\
    \cmidrule(r){2-7} \cmidrule(r){8-13} \cmidrule(r){14-19} 
    SHD & \multirow{2}{*}{insu} & 11 & 1 & 5 & 1 & 0 & \multirow{2}{*}{wate} & 12 & 11 & 5 & 8 & 6 & \multirow{2}{*}{mild} & 8 & 5 & 6 & 6 & 1 \\
F1 & &  .81 &  .97 &  .89 &  .97 &  1.0 &  &  .86 &  .87 &  .94 &  .89 &  .93 &  &  .83 &  .89 &  .87 &  .87 &  .98 \\
\cmidrule(r){1-19}
SHD & \multirow{2}{*}{alar} & 5 & 4 & 1 & 1 & 5 & \multirow{2}{*}{barl} & 13 & 9 & 5 & 8 & 6 & \multirow{2}{*}{hail} & 15 & 15 & 11 & 15 & 13 \\
F1 & &  .93 &  .95 &  .98 &  .98 &  .93 &  &  .88 &  .93 &  .95 &  .92 &  .94 &  &  .74 &  .72 &  .80 &  .72 &  .76 \\
\cmidrule(r){1-19}
SHD & \multirow{2}{*}{hepa} & 8 & 2 & 0 & 1 & 2 & \multirow{2}{*}{win9} & 7 & 1 & 1 & 1 & 1 & \multirow{2}{*}{path} & 1 & 7 & 25 & 7 & 1 \\
F1 & &  .96 &  .99 &  1.0 &  .99 &  .99 &  &  .96 &  .99 &  .99 &  .99 &  .99 &  &  .99 &  .92 &  .77 &  .92 &  .99 \\
\cmidrule(r){1-19}
SHD & \multirow{2}{*}{mun1} & 32 & 7 & 10 & 9 & 15 & \multirow{2}{*}{ande} & 3 & 2 & 0 & 2 & 0 & \multirow{2}{*}{diab} & 18 & 28 & 25 & 26 & 27 \\
F1 & &  .89 &  .98 &  .97 &  .97 &  .95 &  &  .99 &  .99 &  1.0 &  .99 &  1.0 &  &  .97 &  .95 &  .96 &  .96 &  .96 \\
\cmidrule(r){1-19}
SHD & \multirow{2}{*}{pigs} & 0 & 0 & 0 & 0 & 0 & \multirow{2}{*}{link} & 88 & 13 & 0 & 0 & 0 & \multirow{2}{*}{mun} & 107 & 76 & 72 & 93 & 87 \\
F1 & &  1.0 &  1.0 &  1.0 &  1.0 &  1.0 &  &  .93 &  .99 &  1.0 &  1.0 &  1.0 &  &  .93 &  .95 &  .95 &  .94 &  .94 \\
\cmidrule(r){1-19}
SHD & \multirow{2}{*}{mun2} & 117 & 95 & 118 & 110 & 97 & \multirow{2}{*}{mun3} & 151 & 119 & 113 & 99 & 62 & \multirow{2}{*}{mun4} & 165 & 130 & 126 & 146 & 133 \\
F1 & &  .92 &  .93 &  .92 &  .93 &  .93 &  &  .90 &  .92 &  .92 &  .94 &  .96 &  &  .90 &  .92 &  .93 &  .91 &  .93 \\
    \bottomrule
  \end{tabular}
  \end{center}
  \vspace{-1.5em}
\end{table}

\vspace{0.5em}

\noindent\textbf{Tolerance}:
\label{sec:tolerance}
BLIP~\citep{scanagatta2015learning} is the strongest competitor among the three algorithms without skeleton input. To evaluate ML4C's tolerance to imperfect skeletons, we use the skeleton identified by BLIP (i.e., convert its DAG output to the corresponding skeleton) as input for ML4C. The result is shown in Table~\ref{table:tolerance}. Among the 23 datasets, ML4C is better than BLIP on 16 datasets, tied on 4 datasets and BLIP is better than ML4C only on 3 datasets (even for these datasets, ML4C still has very close accuracy to BLIP). Moreover, for the datasets where BLIP produces skeletons with very low accuracy (such as munins, skeleton accuracies$\sim 0.5$), ML4C has significantly better accuracy than BLIP, which shows ML4C's better ability for orientation and also tolerance w.r.t. skeleton misspecification.
\begin{table}[t]
\vspace{-0.4em}
  \tiny
   \setlength{\tabcolsep}{-0.5pt}
   \renewcommand{\arraystretch}{0.2}
  \caption{Tolerance: Use the imperfect skeleton identified by BLIP as ML4C's input. `SA' denotes skeleton accuracy. `M' is short for ML4C. `B' is short for BLIP.}
  \label{table:tolerance}
\begin{center}
\vspace{-2.1em}
  \begin{tabular}{c || c c c c | c c c c | c c c c | c c c c }
    \toprule
     & data & SA & M & B & data & SA & M & B & data & SA & M & B & data & SA & M & B \\
    \cmidrule(r){2-5} \cmidrule(r){6-9} \cmidrule(r){10-13} \cmidrule(r){14-17}
    
SHD & \multirow{2}{*}{asia} & \multirow{2}{*}{ .82} & 6 & 6 & \multirow{2}{*}{canc} & \multirow{2}{*}{ 1.0} & \textbf{ 0 } & 4 & \multirow{2}{*}{eart} & \multirow{2}{*}{ .89} & 5 & 5 & \multirow{2}{*}{sach} & \multirow{2}{*}{ .97} & 14 & \textbf{ 1 }\\
F1 &  & &  .57 &  .57 &  & & \textbf{  1.0 } &  .00 &  & &  .00 &  .00 &  & &  .00 &  .00\\
\cmidrule(r){1-17}
SHD & \multirow{2}{*}{surv} & \multirow{2}{*}{ .91} & \textbf{ 2 } & 6 & \multirow{2}{*}{alar} & \multirow{2}{*}{ .91} & \textbf{ 13 } & 17 & \multirow{2}{*}{barl} & \multirow{2}{*}{ .70} & \textbf{ 52 } & 60 & \multirow{2}{*}{chil} & \multirow{2}{*}{ 1.0} & 0 & 0 \\
F1 &  & & \textbf{  .73 } &  .00 &  & & \textbf{  .84 } &  .82 &  & & \textbf{  .57 } &  .48 &  & &  1.0 &  1.0 \\
\cmidrule(r){1-17}
SHD & \multirow{2}{*}{insu} & \multirow{2}{*}{ .78} & \textbf{ 31 } & 35 & \multirow{2}{*}{mild} & \multirow{2}{*}{ .69} & \textbf{ 31 } & 36 & \multirow{2}{*}{wate} & \multirow{2}{*}{ .48} & \textbf{ 63 } & 65 & \multirow{2}{*}{hail} & \multirow{2}{*}{ .16} & 111 & 111 \\
F1 &  & & \textbf{  .59 } &  .51 &  & & \textbf{  .56 } &  .41 &  & & \textbf{  .25 } &  .20 &  & &  .17 & \textbf{  .18 } \\
\cmidrule(r){1-17}
SHD & \multirow{2}{*}{hepa} & \multirow{2}{*}{ .71} & 85 & \textbf{ 79 } & \multirow{2}{*}{win9} & \multirow{2}{*}{ .71} & \textbf{ 83 } & 103 & \multirow{2}{*}{ande} & \multirow{2}{*}{ .80} & \textbf{ 158 } & 175 & \multirow{2}{*}{diab} & \multirow{2}{*}{ .66} & \textbf{ 522 } & 534 \\
F1 &  & &  .46 & \textbf{  .54 } &  & & \textbf{  .63 } &  .47 &  & & \textbf{  .78 } &  .76 &  & & \textbf{  .44 } &  .43 \\
\cmidrule(r){1-17}
SHD & \multirow{2}{*}{link} & \multirow{2}{*}{ .61} & \textbf{ 916 } & 947 & \multirow{2}{*}{mun1} & \multirow{2}{*}{ .57} & \textbf{ 249 } & 257 & \multirow{2}{*}{path} & \multirow{2}{*}{ .35} & 259 & \textbf{ 241 } & \multirow{2}{*}{pigs} & \multirow{2}{*}{ 1.0} & 12 & \textbf{ 6 } \\
F1 &  & & \textbf{  .53 } &  .49 &  & & \textbf{  .49 } &  .42 &  & & \textbf{  .12 } &  .07 &  & &  .99 & \textbf{  1.0 } \\
\cmidrule(r){1-17}
SHD & \multirow{2}{*}{mun} & \multirow{2}{*}{ .50} & \textbf{ 1484 } & 1599 & \multirow{2}{*}{mun3} & \multirow{2}{*}{ .49} & \textbf{ 1410 } & 1539 & \multirow{2}{*}{mun4} & \multirow{2}{*}{ .45} & \textbf{ 1565 } & 1627 \\
F1 &  & & \textbf{  .36 } &  .29 &  & & \textbf{  .43 } &  .26 &  & & \textbf{  .37 } &  .28 \\
    \bottomrule
  \end{tabular}
  \end{center}
  \vspace{-3em}
\end{table}


\section{Conclusion and Future Work}
\label{sec:conclusion}
We have proposed a supervised causal learning algorithm ML4C, with theoretical guarantee on learnability and remarkable empirical performance. ML4C shows promising results on validating the effectiveness of supervision. To make SCL practical in real-world scenarios, one important direction for future work is to identify the reliable and accurate skeleton from data, considering ML4C requires the skeleton as additional input. Another important future work is
to incorporate continuous data with the identifiability results from various parametric assumptions.\looseness=-1

{\small
\bibliographystyle{siam}
\bibliography{references}}

\begin{thebibliography}{10}

\bibitem{agresti2003categorical}
{\sc A.~Agresti}, {\em Categorical data analysis}, vol.~482, John Wiley \&
  Sons, 2003.

\bibitem{albert2002statistical}
{\sc R.~Albert and A.-L. Barab{\'a}si}, {\em Statistical mechanics of complex
  networks}, Reviews of modern physics, 74 (2002), p.~47.

\bibitem{aliferis2010local}
{\sc C.~F. Aliferis, A.~Statnikov, I.~Tsamardinos, S.~Mani, and X.~D.
  Koutsoukos}, {\em Local causal and markov blanket induction for causal
  discovery and feature selection for classification part i: algorithms and
  empirical evaluation.}, Journal of Machine Learning Research, 11 (2010).

\bibitem{andreassen1987munin}
{\sc S.~Andreassen, M.~Woldbye, B.~Falck, and S.~K. Andersen}, {\em Munin: A
  causal probabilistic network for interpretation of electromyographic
  findings}, in Proceedings of the 10th international joint conference on
  Artificial intelligence-Volume 1, 1987, pp.~366--372.

\bibitem{d2c2015dependency}
{\sc G.~Bontempi and M.~Flauder}, {\em From dependency to causality: a machine
  learning approach.}, J. Mach. Learn. Res., 16 (2015), pp.~2437--2457.

\bibitem{chen2016xgboost}
{\sc T.~Chen and C.~Guestrin}, {\em Xgboost: A scalable tree boosting system},
  in Proceedings of the 22nd acm sigkdd international conference on knowledge
  discovery and data mining, 2016, pp.~785--794.

\bibitem{chickering2002optimal}
{\sc D.~M. Chickering}, {\em Optimal structure identification with greedy
  search}, Journal of machine learning research, 3 (2002), pp.~507--554.

\bibitem{cover1999elements}
{\sc T.~M. Cover}, {\em Elements of information theory}, John Wiley \& Sons,
  1999.

\bibitem{cussens2012bayesian}
{\sc J.~Cussens}, {\em Bayesian network learning with cutting planes}, arXiv
  preprint arXiv:1202.3713,  (2012).

\bibitem{colombo2014stable}
{\sc C.~Diego and H.~M. Marloes}, {\em Order-independent constraint-based
  causal structure learning}, Journal of Machine Learning Research, 15 (2014),
  pp.~3921--3962.

\bibitem{ding2020reliable}
{\sc R.~Ding, Y.~Liu, J.~Tian, Z.~Fu, S.~Han, and D.~Zhang}, {\em Reliable and
  efficient anytime skeleton learning}, in Proceedings of the AAAI Conference
  on Artificial Intelligence, vol.~34, 2020.

\bibitem{erdos1959on}
{\sc P.~Erd{\H{o}}s and R.~A.}, {\em On random graphs}, Publicationes,
  Mathematicae, 6 (1959), pp.~290--297.

\bibitem{fonollosa2019conditional}
{\sc J.~A. Fonollosa}, {\em Conditional distribution variability measures for
  causality detection}, in Cause Effect Pairs in Machine Learning, Springer,
  2019, pp.~339--347.

\bibitem{glymour2019review}
{\sc C.~Glymour, K.~Zhang, and P.~Spirtes}, {\em Review of causal discovery
  methods based on graphical models}, Frontiers in genetics, 10 (2019), p.~524.

\bibitem{gretton2007kernel}
{\sc A.~Gretton, K.~Fukumizu, C.~H. Teo, L.~Song, B.~Sch{\"o}lkopf, A.~J.
  Smola, et~al.}, {\em A kernel statistical test of independence.}, in Nips,
  vol.~20, Citeseer, 2007, pp.~585--592.

\bibitem{hoyer2008nonlinear}
{\sc P.~Hoyer, D.~Janzing, J.~M. Mooij, J.~Peters, and B.~Sch{\"o}lkopf}, {\em
  Nonlinear causal discovery with additive noise models}, Advances in neural
  information processing systems, 21 (2008), pp.~689--696.

\bibitem{janzing2012information}
{\sc D.~Janzing, J.~Mooij, K.~Zhang, J.~Lemeire, J.~Zscheischler,
  P.~Daniu{\v{s}}is, B.~Steudel, and B.~Sch{\"o}lkopf}, {\em
  Information-geometric approach to inferring causal directions}, Artificial
  Intelligence, 182 (2012), pp.~1--31.

\bibitem{kalainathan2019causal}
{\sc D.~Kalainathan and O.~Goudet}, {\em Causal discovery toolbox: Uncover
  causal relationships in python}, arXiv preprint arXiv:1903.02278,  (2019).

\bibitem{koller2009probabilistic}
{\sc D.~Koller and N.~Friedman}, {\em Probabilistic graphical models:
  principles and techniques}, MIT press, 2009.

\bibitem{lam2022greedy}
{\sc W.-Y. Lam, B.~Andrews, and J.~Ramsey}, {\em Greedy relaxations of the
  sparsest permutation algorithm}, in The 38th Conference on Uncertainty in
  Artificial Intelligence, 2022.

\bibitem{lauritzen1996graphical}
{\sc S.~L. Lauritzen}, {\em Graphical models}, vol.~17, Clarendon Press, 1996.

\bibitem{li2020supervised}
{\sc H.~Li, Q.~Xiao, and J.~Tian}, {\em Supervised whole dag causal discovery},
  arXiv preprint arXiv:2006.04697,  (2020).

\bibitem{lopez2015randomized}
{\sc D.~Lopez-Paz, K.~Muandet, and B.~Recht}, {\em The randomized causation
  coefficient.}, J. Mach. Learn. Res., 16 (2015), pp.~2901--2907.

\bibitem{lopezpaz2015towards}
{\sc D.~Lopez-Paz, K.~Muandet, B.~Sch{\"o}lkopf, and I.~Tolstikhin}, {\em
  Towards a learning theory of cause-effect inference}, in Proceedings of the
  32nd International Conference on Machine Learning, PMLR, 2015,
  pp.~1452--1461.

\bibitem{lopez2017discovering}
{\sc D.~Lopez-Paz, R.~Nishihara, S.~Chintala, B.~Scholkopf, and L.~Bottou},
  {\em Discovering causal signals in images}, in Proceedings of the IEEE
  Conference on Computer Vision and Pattern Recognition, 2017, pp.~6979--6987.

\bibitem{lorch2021dibs}
{\sc L.~Lorch, J.~Rothfuss, B.~Sch{\"o}lkopf, and A.~Krause}, {\em Dibs:
  Differentiable bayesian structure learning}, Advances in Neural Information
  Processing Systems,  (2021).

\bibitem{margaritis2003learning}
{\sc D.~Margaritis}, {\em Learning bayesian network model structure from data},
  tech. rep., Carnegie-Mellon Univ Pittsburgh Pa School of Computer Science,
  2003.

\bibitem{meek2013rule}
{\sc C.~Meek}, {\em Causal inference and causal explanation with background
  knowledge}, arXiv preprint arXiv:1302.4972,  (2013).

\bibitem{meek2013strong}
{\sc C.~Meek}, {\em Strong completeness and faithfulness in bayesian networks},
  arXiv preprint arXiv:1302.4973,  (2013).

\bibitem{ng2022masked}
{\sc I.~Ng, S.~Zhu, Z.~Fang, H.~Li, Z.~Chen, and J.~Wang}, {\em Masked
  gradient-based causal structure learning}, in Proceedings of the 2022 SIAM
  International Conference on Data Mining (SDM), 2022.

\bibitem{ramsey2012conservative}
{\sc J.~Ramsey, J.~Zhang, and P.~Spirtes}, {\em Adjacency-faithfulness and
  conservative causal inference}, arXiv preprint arXiv:1206.6843,  (2012).

\bibitem{scanagatta2015learning}
{\sc M.~Scanagatta, C.~P. de~Campos, G.~Corani, and M.~Zaffalon}, {\em Learning
  bayesian networks with thousands of variables.}, in NIPS, 2015,
  pp.~1864--1872.

\bibitem{scutari2012bayesian}
{\sc M.~Scutari}, {\em Bayesian network repository}, URL http://www. bnlearn.
  com,  (2012).

\bibitem{shimizu2006linear}
{\sc S.~Shimizu, P.~O. Hoyer, A.~Hyv{\"a}rinen, A.~Kerminen, and M.~Jordan},
  {\em A linear non-gaussian acyclic model for causal discovery.}, Journal of
  Machine Learning Research, 7 (2006).

\bibitem{smola2007hilbert}
{\sc A.~Smola, A.~Gretton, L.~Song, and B.~Sch{\"o}lkopf}, {\em A hilbert space
  embedding for distributions}, in International Conference on Algorithmic
  Learning Theory, Springer, 2007, pp.~13--31.

\bibitem{PC}
{\sc P.~Spirtes and C.~Glymour}, {\em An algorithm for fast recovery of sparse
  causal graphs}, Social science computer review, 9 (1991), pp.~62--72.

\bibitem{spirtes2000causation}
{\sc P.~Spirtes, C.~N. Glymour, R.~Scheines, and D.~Heckerman}, {\em Causation,
  prediction, and search}, MIT press, 2000.

\bibitem{vijaymeena2016similarity}
{\sc M.~Vijaymeena and K.~Kavitha}, {\em A survey on similarity measures in
  text mining}, Machine Learning and Applications: An International Journal, 3
  (2016), pp.~19--28.

\bibitem{yu2016review}
{\sc K.~Yu, J.~Li, and L.~Liu}, {\em A review on algorithms for
  constraint-based causal discovery}, arXiv preprint arXiv:1611.03977,  (2016).

\bibitem{yu2019dag}
{\sc Y.~Yu, J.~Chen, T.~Gao, and M.~Yu}, {\em Dag-gnn: Dag structure learning
  with graph neural networks}, in International Conference on Machine Learning,
  PMLR, 2019, pp.~7154--7163.

\bibitem{zheng2018dags}
{\sc X.~Zheng, B.~Aragam, P.~Ravikumar, and E.~P. Xing}, {\em Dags with no
  tears: Continuous optimization for structure learning}, arXiv preprint
  arXiv:1803.01422,  (2018).

\end{thebibliography}

\clearpage
\newpage

\begin{appendices}
\numberwithin{equation}{section}

\section{Proofs of Lemmas and Theorems}
\label{app:proofs}
\subsection{Proof of Proposition \ref{prop:iden_rand_guess}}
\IDENRANDGUESS*
We take learning target as the orientation of an edge as an example, so we are analyzing the performance of a binary classifier against random guessing. The conclusion can be easily extended to general case.

Denote random guessing as a degenerated estimator $r(X)\equiv 0.5$, which indicates the probability of label = 1 is always 0.5, regardless of any input.

Denote the joint probability distribution of $X$ and $Y$ as $\mathbb{P}$ and the space of all joint probability distribution is $\mathcal{P}$, then we aim to prove the following statement which is in an adversarial setting:
\vspace{0.5em}
\resizebox{\hsize}{!}{
$r=\argmin_{f \in \mathcal C}\mathop{\sup }_{\mathbb{P} \in \mathcal{P}}E_{ \left( X,Y \right)  \sim \mathbb{P}} \left[ -Y\log f \left( X \right) - \left( 1-Y \right) \log  \left( 1-f \left( X \right)  \right)  \right]$
}
\vspace{0.3em}
The expectation is the standard binary cross entropy loss; we are allowed to enumerate every possible joint probability distribution in $\mathcal{P}$ because the learning target is non-identifiable. $\mathcal{C}$ is the space of all possible binary classifiers.

\begin{proof}
Given any binary classifier $f$, we partition the space of $X$ by $A$, $B$ and $C$ where $A=\{x\vert f(x)>0.5\}$, $B=\{x\vert f(x)<0.5\}$, $C=\{x\vert f(x)=0.5\}$. Then we construct the following joint probability distribution $P^{\ast}$:
$$P^{\ast} \left( X,Y \right) = \left\{ \begin{matrix}
P^{\ast} \left( Y=0 \vert X=x \right) =1 \ \text{if} \ x \in A\\
P^{\ast} \left( Y=1 \vert X=x \right) =1 \ \text{if} \ x \in B\\
\text{arbitrary} \ \text{if} \ x \in C\\
\end{matrix}
 \right\}$$
Then it is easy to see that \( E_{ \left( X,Y \right)  \sim P^{\ast}} \left[ -Y\log f \left( X \right) - \left( 1-Y \right) \log  \left( 1-f \left( X \right)  \right)  \right] \geq 1 \). Note that \( E \left[ -Y\log r \left( X \right) - \left( 1-Y \right) \log  \left( 1-r \left( X \right)  \right)  \right]  \equiv 1 \), thus \( r \) achieves minimum worse-case loss.
\end{proof}



\subsection{Proof of Lemma \ref{lem:sepsets_nonempty}}
\SEPSETSNONEMPTY*

\begin{proof}
According to Lemma 3.3.9 of ~\citep{spirtes2000causation}, in a directed acyclic graph $G$, if $X$ is not a descendant of $Y$, and $X$ and $Y$ are not adjacent, then $X$ and $Y$ are d-separated by \textbf{Parents}$(Y)$. Given an UT $\UTXTY$, $X$ and $Y$ are not adjacent. Either $X$ is not a descendant of $Y$, or $Y$ is not a descendant of $X$, otherwise a loop will be introduced. Thus there either exists \textbf{Parents}$(X) \equiv PC_X\cup T$, or \textbf{Parents}$(Y) \equiv PC_Y\cup T$, which belongs to $\Scal$. Thus $\Scal$ is non-empty.
\end{proof}

\subsection{Proof of Lemma \ref{lem:existence_of_strong_discriminative_predicate}}
\STRONGEXISTS*

\begin{proof}
\label{prf:lemma_strong}
According to Theorem 5.1 on p.410 of~\citep{spirtes2000causation}, for a canonical dataset with infinite samples (i.e., assuming faithfulness and perfect conditional independence information), if an UT $\UTXTY$ is a v-structure, then $T$ does not belong to any separation sets of $(X,Y)$; if an UT $\UTXTY$ is not a v-structure, then $T$ belongs to every separation sets of $(X,Y)$. 

Note that in our setting (a canonical dataset with infinite samples), there is no ambiguity: T is either in all the separation sets or none of them. Now we use this statement as a tool for our proof.
\begin{enumerate}
    \item Predicate $\olp(T,\Scal)=0\Longleftrightarrow \forall S\in\Scal , T\notin S$, which states that predicate is $\true$ if and only if $T$ is not in any separation set of $X$ and $Y$, i.e., $\UTXTY$ is a v-structure. Hence the predicate is a strong predicate.
    \item Predicate $\olp(T,\Scal)<0.5$ indicates that only if more than half of the separation sets do not contain $T$, then the UT is oriented as a v-structure. Since there is no ambiguity in our setting (i.e., $\olp(T,\Scal)$ can only be either $0$ or $1$), $\olp(T,\Scal)<0.5 \Leftrightarrow \olp(T,\Scal)=0$. 
    \item Predicate $\olp(T,\Scal)<1\wedge\min\{X\sim Y\vert T\cup \Scal\}>0$ is composed of two parts. Since there is no no ambiguity in our setting (i.e., $\olp(T,\Scal)$ can only be either $0$ or $1$), $\olp(T,\Scal)<1 \Leftrightarrow \olp(T,\Scal)=0$. It is shown that $\min\{X\sim Y\vert T\cup \Scal\}>0$ is redundant given $\olp(T,\Scal)<1$: $\olp(T,\Scal)<1 \Rightarrow \olp(T,\Scal)=0 \Rightarrow \UTXTY$ is a v-structure$\Rightarrow \min\{X\sim Y\vert T\cup \Scal\}>0$, since conditioning on $T$ unblocks the path $X-T-Y$. Thus, $\olp(T,\Scal)<1\wedge\min\{X\sim Y\vert T\cup \Scal\}>0 \Leftrightarrow \olp(T,\Scal)<1 \Leftrightarrow \olp(T,\Scal)=0$. 
\end{enumerate}
\end{proof}

\subsection{Proof of Lemma \ref{lem:existence_of_weak_discriminative_predicate}}
\WEAKEXISTS*

\begin{proof}
For a canonical dataset with infinite samples,
\begin{enumerate}
	\item \textbf{$\left\{ X \sim Y \vert T \right\} >0$}: \begin{inparaenum}[1)]\item $\UTXTY$ is a v-structure $\Rightarrow$ $T$ is a collider $\Rightarrow$ $T$ unblocks $X$ and $Y$ through path $X-T-Y$ $\Rightarrow$ $\left\{ X \sim Y \vert T \right\} >0$ holds $\true$. \item if $\UTXTY$ is not a v-structure, then $\left\{ X \sim Y \vert T \right\} >0$ can be $\true$ or $\false$. e.g., it is $\false$ for $X\rightarrow T \rightarrow Y$ (no more paths connect $X$ and $Y$), but if there exists another node $X\rightarrow T^{'} \rightarrow Y$, it is evaluated $\true$. \end{inparaenum}  Therefore, it satisfies criterion ii) of definition \ref{def:weak_strong_discriminative_predicate}, but not i) hence it is a weak discriminative predicate.

	\item \textbf{$\left\{ X \sim Y \vert \mathcal{PC}_{T} \right\} =0$}: \begin{inparaenum}[1)]\item $\UTXTY$ is not a v-structure $\Rightarrow$ $T$ is a non-collider $\Rightarrow$ $\forall pc_t\in \mathcal{PC}_{T}$, there exists a path $X-T-Y$ from $X$ to $Y$, where $T$ is the only node on path, $T$ is a non-collider, and $T\notin \{pc_t\}$ $\Rightarrow$ $pc_t$ does not block the path $\Rightarrow$ $\{ X \sim Y \vert \mathcal{PC}_{T} \} =0$ always holds $\false$. \item if $\UTXTY$ is a v-structure, then $\left\{ X \sim Y \vert \mathcal{PC}_{T} \right\} =0$ can be $\true$ or $\false$. \end{inparaenum} Therefore, it satisfies criterion i) but not ii) hence it's a weak discriminative predicate.

	\item \textbf{$\left\{ PC_{X} \sim PC_{Y} \vert \Scal \cup T \right\} >0$}: \begin{inparaenum}[1)]\item $\UTXTY$ is a v-structure $\Rightarrow$ $T$ is a collider $\Rightarrow$ $\forall pc_x\in PC_{X}, pc_y\in PC_{Y}, S\in\mathcal{S}$, $S\cup T$ unblock $pc_x$ and $pc_y$ through path $pc_x-X-T-Y-pc_y$ $\Rightarrow$ $\left\{ pc_x \sim pc_y \vert S \cup T \right\} >0$ always hold $\true$. \item if $\UTXTY$ is not a v-structure then it can be $\true$ or $\false$. \end{inparaenum}  Therefore, it satisfies criterion ii) but not i) hence it's a weak discriminative predicate.
\end{enumerate}
\end{proof}

\subsection{Proof of Theorem \ref{thrm:tend_perfect}}
\TENDPERFECT*

\begin{proof}
According to Lemma \ref{lem:existence_of_strong_discriminative_predicate}, there exists strong discriminative predicate $P$ which achieves zero loss given a canonical dataset and sufficient samples. Thus, when adequate ML model is chosen, ML4C-Learner can achieve no worse performance than $P$ (e.g., we can set the parameters of ML4C-Learner so that it approximates predicate $P$ initially, and then apply standard gradient descent procedure).
\end{proof}

\section{Implementation Details}
\label{app:implementation}
\subsection{Calculating conditional dependencies}
There are several ways to measure the conditional dependence, such as p-value by testing of conditional independence, or conditional mutual information~\citep{cover1999elements}. For categorical variables, a good choice is $G^2$ test~\citep{agresti2003categorical}. In our implementation, we adopt an approximate version of $G^2$ statistic, and use p-value to measure the conditional dependence.

Moreover, considering p-value can easily vanish due to numerical precision in 64-bit computers. Therefore, we use a transformation of p-value to avoid the issue, as additional quantity to measure conditional dependency. We first define complementary error function as$$g \left( z \right) =1-\frac{2}{\sqrt[]{ \pi }} \int _{0}^{z}e^{-t^{2}}\mathrm{d}t,$$ and we use quantity $z$ by inverse of $g$: $$z=g^{-1}(x).$$

Given a p-value $x$, we use $g^{-1}$ as a non-linear transformation to obtain a better re-scaled quantity to measure conditional dependency. Intuitively, $z$ can be viewed as $z$-sigma for a standard normal distribution, e.g., if p-value is 0.05, then $z=2$, since 2-sigma indicates probability of values that lie within 2-sigma interval in a normal distribution is 0.95.

\subsection{ML4C Training and inference details}
\subsubsection{Data synthesis details}

\textbf{Graph structure:} We adopt the Erd\H{o}s-R\'{e}nyi (ER) model~\citep{erdos1959on} and the Scale-Free (SF) model~\citep{albert2002statistical}, which are two commonly used model for graph synthesis. We categorize the scale of the graph (number of nodes $d$) into four classes: small, medium, large, and very large, corresponding to $d$ being uniformly sampled from intervals $[10,20]$, $[21,50]$, $[51,100]$, and $[101,1000]$, respectively. Given the number of nodes $d$, the sparsity of the graph (defined as the ratio of the average number of edges to the number of nodes, i.e., the average in-degree of all nodes) is randomly sampled from a uniform distribution $[1.2, 1.7]$. Given the number of nodes and the expected number of edges, the graph skeleton is generated accordingly by the two random graph models. Then the skeleton is randomly oriented to a DAG by upper triangular permutation. 

\noindent\textbf{Conditional probability table:} Now we illustrate how we come up with Conditional Probability Table (CPT) for each node. In accordance with the topological ordering of the graph, each node is first assigned its cardinality, which is randomly sampled from a truncated normal distribution $\mathcal{N}(\mu=2, \sigma=\frac{1.5}{m}, \min=2)$, where $m$ denotes the maximum number of peers of the node (i.e. $\max\{ \text{in-degree of the effect nodes of this node}\}$). This regularization is designed to make the forward sampling process faster and prevent some certain nodes with many cause nodes from getting stuck. Since the number of different conditions to be enumerated is exponential ($\Pi_{c \in \text{ causes}}{\text{cardinality}_c}$), node with a larger maximum peers number tends to have smaller cardinality. Next, we enumerate each of its unique conditions (given by combinations of its cause nodes' cardinalities) and randomly generate its probability distribution at each condition. The probability distribution is sampled from a Dirichlet distribution with parameter $\alpha\sim U[0.1, 1.0]$ and grid number as this code's cardinality.

\noindent\textbf{Training data:} Having CPT specification of each node, a sample of 10k rows of observations is obtained for each graph according to the standard Bayesian network forward sampling. This generates a total dataset of 4 scales $\times$ 2 graph models $\times$ 50 graphs for each class = 400 unique graphs and the corresponding sampled data. Different SCL algorithms are then further used to extract the required features corresponding to the respective learning targets, e.g., all edges of all graphs for pairwise SCL algorithms. For our ML4C learning targets, all UTs are extracted from graphs, consisting of a total of 97,010 V-structures (label=1) and 195,691 non-V-structures (label=0).

\noindent\textbf{Kernel mean embedding:} Dimensionality of feature vector is 755. Specifically, to represent each extended conditional dependency $D=\mathbf{A}\sim\mathbf{B}\vert \mathcal{Z}$ (i.e., a set of scalars with varied set size, line 190), we use standard kernel mean embedding technique in \citep{smola2007hilbert} to obtain the embedded vector $ \frac { 1} { | D | } \sum _ { z \in D } \left( \cos \left( \left\langle w _ { j } , z \right\rangle + b _ { j } \right) \right) _ { j = 1 } ^ { m } \in \mathbb { R } ^ { m }$. Here m=15, which means embedding dimensionality for each extended conditional dependency is 15. We also include 5 additional statistics (max, min, mean, std, and set size). Besides, we use both p-value and severity to quantify each extended conditional dependency separately. Of the 4 bivariable $\times$ 5 conditional = 20 pairs, only one $(X;Y|T)$ is unitary (i.e., it is a single scalar), so we apply embedding to the rest 19 pairs. In addition, entanglement is considered in terms of 5 scaling and 7 overlap coefficients. Thus, the total dimensionality of feature vector is 755 = 5 (scaling) + 7 (overlaps) + 2 * 1 (unitary) + [1 (set size) + 2 * 4 (mean/std/max/min) + 2 * 15 (embedding dimensionality)] * 19 (pairs). See our code for more details at implementation level (Tools/Utility.py L61, BayesianNetwork/CITester.py L104, Experiments/GenerateFeatures.py L121).

\subsubsection{XGBoost hyper-parameter settings}
We use \texttt{xgb.XGBClassifier()}, the Python API provided by XGBoost~\citep{chen2016xgboost}, to implement the binary classifier ML4C-Learner. All hyper-parameters are set as default. We set the threshold value $T$ = 0.1.


\subsection{Post processing}

Although ML4C-Learner achieves high accuracy on classifying UTs into v-structures or non-v-structures (UT-F1 = 0.9, as shown in Table~\ref{table:accuracy}), it is still possible to have conflicts among the detected v-structures. We adopt a straightforward heuristic to resolve conflicts: suppose we have two conflict v-structures $A \rightarrow B \leftarrow C$ and $B \rightarrow C \leftarrow D$, we discard the one with lower probability score (by ML4C-Learner). We continue such pairwise conflict resolving until no more conflicts exist. We use the left v-structures to construct the partial DAG (bottom-right of Figure~\ref{fig:two_phase_workflow}(b)). Pseudo-code is shown in Algorithm~\ref{alg:conflict}.


\begin{algorithm} [htbp] 
\caption{Conflict resolving} 
\label{alg:conflict}
\SetKwFor{For}{for}{do}{}
\SetKwFor{If}{if}{then}{}
\SetKwInOut{Input}{input} \SetKwInOut{Output}{output}
\Input{ v-structure candidates $VC=\{v_1, \cdots, v_p\}$,\\
        score querier $s: v_i \rightarrow s_i$, returning $v_i$'s probability score\\ }
\Output{ Final v-structure candidates $FV$, which is self-consistent. }
\textbf{Initialize:} removing v-structure set $RV$.\\
\For{ $v_i \in VC $}{
    $s_i\leftarrow s(v_i)$\\
    flag$\leftarrow\false$\\
    \For{ $v_j\in VC$ }{
        $s_j\leftarrow s(v_j)$\\
        \If{$v_i$\normalfont{ conflicts with }$v_j$ \textbf{and} $s_i<s_j$}{
            flag$\leftarrow\true$\\
            \textbf{break}
        }
    }
    \If{\normalfont{flag}}{
        SV $\leftarrow$ SV$\cup\{v_i\}$
    }
}
FV$\leftarrow$VC$\backslash$RV.
\end{algorithm}

\begin{table}[!b]
\scriptsize
  \caption{SHD calculation details.}
  \label{table:metrics}
  \begin{center}
  \begin{tabular}{|c|c|c|c|c|}
  \hline
    \multirow{2}{*}{\makecell{in result CPDAG→\\in truth CPDAG↓}} & \multicolumn{2}{c|}{iden (directed)} & \multirow{2}{*}{\makecell{uniden\\(undirected)}} & \multirow{2}{*}{\makecell{missing in\\skeleton}} \\ \cline{2-3}
    & right & wrong  &  & \\ \hline
iden & \cmark \circled{1}  & \xmark \circled{2} & \xmark\circled{3}  & \xmark\circled{4} \\ \hline
uniden & \multicolumn{2}{c|}{\xmark\circled{5}} & \cmark \circled{6}  & \xmark\circled{7} \\ \hline
nonexist  & \multicolumn{2}{c|}{\xmark\circled{8}} & \xmark\circled{9} & \cmark \circled{10} \\ \hline
\end{tabular}
\end{center}
\end{table}

\subsection{Reproducibility}
Our performance result is highly reproducible. Our classifier is a vanilla XBGoost classifier with fixed hyper-parameters, thus its output is fully deterministic given a specific input. Our featurization follows deterministic process, as depicted in~\cref{sec:featurization}. The only uncertainty comes from kernel mean embedding, where we follow standard approach without any modification. We observed that its impact on perturbation of result is negligible. Please check our code for details of reproducibility.

\section{Details of Evaluation}
\label{app:evaluation}
\begin{table*}[t]
  \scriptsize
   \setlength{\tabcolsep}{3.9pt}
  \caption{Transferability: ML4C trains/tests both on synthetic datasets with different configurations.}
  \label{table:transferability}

\begin{center}
\vspace{-1em}
  \begin{tabular}{c  c || c c c  | c c c | c c c  | c c c }
    \toprule
       & train & test & SHD & F1 & test & SHD & F1 & test & SHD & F1 & test & SHD & F1 \\
    \midrule
\multirow{4}{*}{ \rotatebox[origin=c]{90}{\# node} }
& 10 & 10 &  1.2$\pm$2.4 &  .94$\pm$.12 & 50 &  4.8$\pm$3.4 &  .95$\pm$.03 & 100 &  6.6$\pm$4.7 &  .97$\pm$.02 & 1k &  50.6$\pm$8.4 &  .97$\pm$.00 \\
& 50 & 10 &  0.4$\pm$0.8 &  .97$\pm$.05 & 50 &  0.8$\pm$1.0 &  .99$\pm$.01 & 100 &  4.4$\pm$4.7 &  .98$\pm$.02 & 1k &  23.2$\pm$5.7 &  .99$\pm$.00 \\
& 100 & 10 &  0.0$\pm$0.0 &  1.0$\pm$.00 & 50 &  1.2$\pm$1.6 &  .99$\pm$.01 & 100 &  4.0$\pm$4.6 &  .98$\pm$.02 & 1k &  21.6$\pm$4.8 &  .99$\pm$.00 \\
& 1k & 10 &  0.4$\pm$0.8 &  .97$\pm$.05 & 50 &  0.8$\pm$1.0 &  .99$\pm$.01 & 100 &  1.4$\pm$2.3 &  .99$\pm$.01 & 1k &  14.8$\pm$8.2 &  .99$\pm$.00 \\
\cmidrule(r){1-14}
\multirow{4}{*}{ \rotatebox[origin=c]{90}{sparsity} }
& 1 & 1 &  0.8$\pm$1.6 &  .99$\pm$.02 & 2 &  3.4$\pm$2.9 &  .97$\pm$.02 & 3 &  3.0$\pm$2.5 &  .98$\pm$.01 & 4 &  11.4$\pm$3.9 &  .95$\pm$.02 \\
& 2 & 1 &  1.8$\pm$1.6 &  .98$\pm$.02 & 2 &  2.2$\pm$1.7 &  .98$\pm$.01 & 3 &  2.2$\pm$2.0 &  .99$\pm$.01 & 4 &  8.2$\pm$2.5 &  .97$\pm$.01 \\
& 3 & 1 &  1.0$\pm$1.3 &  .98$\pm$.02 & 2 &  2.2$\pm$1.3 &  .98$\pm$.01 & 3 &  4.4$\pm$3.6 &  .97$\pm$.02 & 4 &  4.0$\pm$3.2 &  .98$\pm$.01 \\
& 4 & 1 &  2.4$\pm$2.3 &  .97$\pm$.03 & 2 &  2.2$\pm$1.9 &  .98$\pm$.01 & 3 &  3.2$\pm$2.7 &  .98$\pm$.02 & 4 &  4.8$\pm$3.7 &  .98$\pm$.01 \\
\cmidrule(r){1-14}
\multirow{4}{*}{ \rotatebox[origin=c]{90}{samplesize} }
& 1k & 1k &  2.8$\pm$2.3 &  .97$\pm$.02 & 5k &  2.0$\pm$2.2 &  .98$\pm$.02 & 10k &  1.6$\pm$2.3 &  .98$\pm$.02 & 20k &  1.0$\pm$1.3 &  .99$\pm$.01 \\
& 5k & 1k &  5.2$\pm$2.9 &  .95$\pm$.03 & 5k &  1.0$\pm$2.0 &  .99$\pm$.02 & 10k &  2.2$\pm$3.5 &  .98$\pm$.04 & 20k &  0.6$\pm$0.8 &  .99$\pm$.01 \\
& 10k & 1k &  5.2$\pm$4.8 &  .95$\pm$.05 & 5k &  1.8$\pm$2.7 &  .98$\pm$.02 & 10k &  2.0$\pm$3.1 &  .98$\pm$.03 & 20k &  0.6$\pm$0.8 &  .99$\pm$.01 \\
& 20k & 1k &  4.8$\pm$3.3 &  .95$\pm$.03 & 5k &  2.4$\pm$2.6 &  .98$\pm$.02 & 10k &  1.2$\pm$1.6 &  .99$\pm$.02 & 20k &  1.0$\pm$1.3 &  .99$\pm$.02 \\
\cmidrule(r){1-14}
\multirow{2}{*}{ \rotatebox[origin=c]{90}{gtype} }
& ER & ER &  1.0$\pm$2.0 &  .99$\pm$.02 & SF &  2.2$\pm$1.6 &  .98$\pm$.01 & \multicolumn{6}{c}{} \\
& SF & ER &  1.6$\pm$1.9 &  .98$\pm$.02 & SF &  2.2$\pm$2.4 &  .98$\pm$.02 & \multicolumn{6}{c}{} \\
    \bottomrule
  \end{tabular}
  \end{center}

  \vspace{-1em}
\end{table*}
\subsection{Evaluation metrics}

We calculate SHD at CPDAG level. Specifically, SHD is computed between the learned CPDAG$(\hat{G})$ and ground truth CPDAG$(G)$, i.e., the smallest number of edge additions, deletions, direction reversals and type changes (directed vs. undirected) to convert the output CPDAG to ground truth CPDAG. As is shown in Table~\ref{table:metrics}, SHD is equal to the sum of the number of \xmark s in the table.

F1-score is then calculated based on the identifiable edges of CPDAG$(\hat{G})$ and CPDAG$(G)$, where the accuracy (precision) is equal to True Positive Rate (TPR) and the recall (recall) is equal to 1 - False Discovery Rate (FDR). Details about the specific calculation can also refer to Table~\ref{table:metrics}:
$$\text{precision=TPR}=\frac{\circled{1}}{\circled{1}+\circled{2}+\circled{3}+\circled{4}},$$
$$\text{recall=1-FDR}=\frac{\circled{1}}{\circled{1}+\circled{2}+\circled{5}+\circled{8}},$$



\subsection{Predicates in Table~\ref{table:reliability}: Reliability}
Table~\ref{table:reliability} shows the performance of 4 weak discriminative predicates and 4 strong discriminative predicates. Specifically, the four strong predicates are respectively 1) $t\sim U[0,1],\ \olp(T, \Scal)\geq t$; 2) $\olp(T, \Scal)==0$; 3) $\olp(T, \Scal)==0$ and $\{X\sim Y\vert \Scal\cup T\}>0$; 4) $\{X\sim Y\vert \Scal\vee T\}>0$. The four weak predicates are respectively 1) $\{PC_X\sim PC_Y\vert T\}>0$; 2) $\{PC_X\sim PC_Y\vert \Scal\vee T\}>0$; 3) $\{X\sim Y\vert PC_T\}==0$; 4) $\{X\sim Y\vert \Scal \vee \mathcal{PC}_T\}>0$.

\subsection{Table~\ref{table:transferability}: ML4C's Transferability}
To evaluate whether ML4C generalizes well to various types of testing data, we vary scale (\#nodes), graph sparsity, generating mechanism and sample size. ML4C is trained on a fixed configuration but it is tested with different domains (i.e., data generated under different configuration). ML4C transfers well on different domains, e.g., even if it is trained on 10 nodes but tested on 1,000 nodes, the F1-score only drops 0.02.


When we test transferability over one domain (e.g., the first bar, \#nodes), then \#nodes is set from 4 options (10, 50, 100, 1k), and $4\times 4=16$ pairs of train-test experiments are conducted. For each experiment, 50 graphs are synthesized for training and another 5 graphs for test. Except for the target domain (\#nodes), all the other domains use the default configuration. The result SHD and F1-score are reported as mean value and standard deviation over the five test graphs.%

\section{Code and Data}
\label{app:codedata}
\subsection{URLs of all competitors}
We use open-source codes of other algorithms for evaluation.

For Jarfo, RCC, NCC, GES, GS(Grow-Shrink), and CDS, we use the API provided by Causal Discovery Toolbox~\citep{kalainathan2019causal}: \url{https://github.com/FenTechSolutions/CausalDiscoveryToolbox}.

For HC(Hill-Climbing) we use pgmpy \url{https://github.com/pgmpy/pgmpy} with BDeu score.

For PC we use the R package pcalg \url{https://cran.r-project.org/web/packages/pcalg}.

For Conservative-PC and Majority-rule PC, we slightly modify the source code of pcalg to enable a faster run on large scale datasets. GLL-MB is also implemented based on pcalg.

For GOBNILP we use \url{https://bitbucket.org/jamescussens/pygobnilp/}.

\subsection{Algorithms starting from data: DAG-GNN/BLIP/GRaSP}
\subsubsection{Code URL}
\begin{enumerate}
    \item BLIP: \url{https://cran.r-project.org/web/packages/r.blip/}.
    \item DAG-GNN: We use a clean Python implementation of the DAG-GNN algorithm: \url{https://github.com/ronikobrosly/DAG_from_GNN/}.
    \item GRaSP: \url{https://github.com/cmu-phil/causal-learn/blob/main/causallearn/search/PermutationBased/GRaSP.py}. Actually GRaSP just had a lastest release in Tetrad java GUI that supports skeleton as input. We did not use that since it is difficult to specify large graphs in GUI.
\end{enumerate}
\subsubsection{Hyper-parameter settings}
\begin{enumerate}
    \item Time limit: The running time of all programs is limited to 24 hours.
    \item Max-in-degree: The max-in-degree threshold for BILP is set to 6. The max-in-degree threshold for GOBNILP is set to 3.
    \item Configurations of DAG-GNN are as follows. Epochs=300, batch size=100, learning rate=3e-3, graph threshold=0.3. Graph threshold is a threshold for weighted adjacency matrix (i.e., any weights $> -0.3$ and $< 0.3$ means the two variables are not adjacent).
\end{enumerate}
\subsubsection{Verifying the results of DAG-GNN}
To make sure DAG-GNN is correctly executed, we have carefully experiment DAG-GNN from the following two aspects:
\paragraph{Reproducing the results of paper~\citep{yu2019dag}}
We take the child dataset as an example to test the reproducibility, because the data set has been reported by ~\citep{yu2019dag}.
\begin{table}[h]
  \small
   \caption{Reproducing results for child}
   \label{reprochild}
\begin{center}
  \begin{tabular}{c c c}
    \toprule
    & groundtruth & child \\
    \midrule
    BIC & -1.23e+4 & -1.36e+4 \\
    \bottomrule
  \end{tabular}
  \end{center}
  \vspace{-1em}
\end{table}
As can be seen from Table~\ref{reprochild}, the BIC scores are similar to the results reported in the original paper (child: -1.38e+4). That is to say, the results in Yu et al.’s paper are reproduced by us.
\paragraph{Different graph thresholds}
 The following are the BIC scores on the data sets of alarm and water with different graph thresholds.
\begin{table}[h]
  \scriptsize
  \setlength{\tabcolsep}{3pt}
   \caption{Results with different graph thresholds}
   \label{graphthresholds}
\begin{center}
  \begin{tabular}{c c c c c c c}
    \toprule
    & Groundtruth & 0.1 & 0.2 & 0.3 & 0.4 & 0.5 \\
    \midrule
    alarm & -1.08e+5 & -1.90e+5 & -1.44e+5 & -1.59e+5 & -1.77e+5 & -1.91e+5\\
    water & -1.35e+5 & -1.32e+5 & -1.37e+5 & -1.44e+5 & -1.53e+5 & -1.62e+5 \\
    \bottomrule
  \end{tabular}
  \end{center}
  \vspace{-1em}
\end{table}
The graph threshold recommended by \citep{yu2019dag} is 0.3. It can be seen that the performance is stable when the threshold is around 0.3. We have verified that there are similar conclusions on other data sets. Therefore 0.3 should be a reasonable threshold.

\section{Discussions}
\label{app:discussions}
\subsection{Causal discovery on discrete data vs. continuous data}
In our work, we target the problem of causal discovery from discrete data. Operationally, our approach can be easily extended to work on continuous data since there are standard algorithms in literature (e.g., Hilbert-Schmidt Independence Criterion~\citep{gretton2007kernel}) to be used to calculate conditional dependencies over numerical variables. However, the theory of causal structure identifiability on continuous data can be different. According to the theorem of Markov completeness~\citep{meek2013strong}, for discrete data, we can only identify a causal graph up to its Markov equivalence class (i.e., represented by CPDAG). But for continuous data, if it further satisfies additive noise model~\citep{hoyer2008nonlinear}, or linear SEM with non-Gaussian noise~\citep{shimizu2006linear}, some undirected edges in the CPDAG can further be oriented. As a result, on continuous data, the two learning phases (in the two-phase learning paradigm which we have advocated) may not be easily facilitated by a single learning task (i.e., v-structures are no longer sufficient for identifiability), and there's needs to design the corresponding learning tasks with more considerations. In summary, we believe the two-phase learning paradigm sheds light on fundamental question of SCL, it shows an appropriate aspect to conduct SCL on continuous data, which will be our future work.

\subsection{Training Data for Supervised Causal Learning}
In our experiment, we use mild scale of training set: 400 graphs, with ~100,000 v-structures (label = 1) and ~300,000 non-v-structures (label = 0) and we already can achieve good results (note: each training instance here refers to a dataset with ground-truth DAG as label).
An interesting perspective is, supervised causal learning is beneficial on that obtaining training set is cheap and straightforward since we can sample graphs from DAG space and generate corresponding datasets. Therefore, we could generate as many training instances as we want to push the learner's performance. We believe this is a strong support on promoting supervised causal learning as one promising direction in causal discovery.

\end{appendices}

\end{document}